%% file: main.tex
\documentclass[runningheads]{llncs}

\usepackage{eccv}
\usepackage{eccvabbrv}
\usepackage{graphicx}
\usepackage{booktabs}
\usepackage{colortbl}
\usepackage{wrapfig}
\usepackage{float}
\usepackage{multirow}
\usepackage{pifont}
\usepackage{algorithm2e}
\usepackage{algorithmic}
\usepackage{subcaption}
\usepackage{adjustbox}
\usepackage[accsupp]{axessibility}  

\usepackage{hyperref}

\usepackage{orcidlink}

\newcommand{\myparagraph}[1]{\vspace{3pt}\noindent{\bf #1}}

\begin{document}

\title{VisionTrap: Vision-Augmented Trajectory Prediction Guided by Textual Descriptions}
\titlerunning{VisionTrap}

\author{Seokha Moon\inst{1}\orcidlink{0009-0009-0506-5958} \and
Hyun Woo\inst{1}\orcidlink{0009-0005-5217-6379} \and
Hongbeen Park\inst{1}\orcidlink{0009-0003-2633-288X} \and
Haeji Jung\inst{1}\orcidlink{0009-0008-8347-7432} \and \\
Reza Mahjourian\inst{2}\orcidlink{0000-0002-4457-8395} \and 
Hyung-gun Chi\inst{3}\orcidlink{0000-0001-5454-3404} \and
Hyerin Lim\inst{4}\orcidlink{0009-0003-3369-8169} \and 
Sangpil Kim\inst{1}\orcidlink{0000-0002-7349-0018} \and  
Jinkyu Kim\inst{1}\orcidlink{0000-0001-6520-2074}}

\authorrunning{S. Moon et al.}
\institute{
Korea University, Seoul 02841, Republic of Korea \and
The University of Texas at Austin, Texas 78712, USA \and
Perdue University, West Lafayette 95008, USA \and
Hyundai Motor Company, Seongnam 13529, Republic of Korea \\
}

\maketitle

\def\thefootnote{*}\footnotetext{Corresponding author: J. Kim (jinkyukim@korea.ac.kr)}

\input{./sec/0_abstract}    
\input{./sec/1_introduction}
\input{./sec/2_relatedwork}
\input{./sec/3_method}
\input{./sec/4_dataset}

\input{./sec/5_experiment}
\input{./sec/6_conclusion}
\subsubsection*{Acknowledgment.}

\small
This work was supported by Autonomous Driving Center, Hyundai Motor Company R\&D Division. This work was partly supported by IITP under the Leading Generative AI Human Resources Development(IITP-2024-RS-2024-00397085, 10\%) grant, IITP grant (RS-2022-II220043, Adaptive Personality for Intelligent Agents, 10\% and IITP-2024-2020-0-01819, ICT Creative Consilience program, 5\%). This work was also partly supported by Basic Science Research Program through the NRF funded by the Ministry of Education(NRF-2021R1A6A1A13044830, 10\%). This work also supported by Culture, Sports and Tourism R\&D Program through the Korea Creative Content Agency grant funded by the Ministry of Culture, Sports and Tourism in 2024((International Collaborative Research and Global Talent Development for the Development of Copyright Management and Protection Technologies for Generative AI, RS-2024-00345025, 4\%),(Research on neural watermark technology for copyright protection of generative AI 3D content, RS-2024-00348469, 25\%)), 
Institute of Information \& communications Technology Planning \& Evaluation (IITP) grant funded by the Korea government(MSIT)(RS-2019-II190079, 1\%). We also thank Yujin Jeong and Daewon Chae for their helpful discussions and feedback.
%
%
\bibliographystyle{splncs04}
\bibliography{main}
\clearpage

\title{Supplemental Material}
\titlerunning{VisionTrap}
\author{}
\authorrunning{S. Moon et al.}
\institute{}
\maketitle

\section{Details for Evaluation and Implementation}
\input{./src_supple/sec/setup}

\section{More Detail for nuScenes-Text Dataset}
\input{./src_supple/sec/dataset}

\section{Further Results}
\input{./src_supple/sec/result}

\end{document}

%% file: sec/0_abstract.tex
\begin{abstract}
Predicting future trajectories for other road agents is an essential task for autonomous vehicles.  Established trajectory prediction methods primarily use agent tracks generated by a detection and tracking system and HD map as inputs.
In this work, we propose a novel method that also incorporates visual input from surround-view cameras, allowing the model to utilize visual cues such as human gazes and gestures, road conditions, vehicle turn signals, etc, which are typically hidden from the model in prior methods. Furthermore, we use textual descriptions generated by a Vision-Language Model (VLM) and refined by a Large Language Model (LLM) as supervision during training to guide the model on what to learn from the input data.
Despite using these extra inputs, our method achieves a latency of 53 ms, making it feasible for real-time processing, which is significantly faster than that of previous single-agent prediction methods with similar performance.
Our experiments show that both the visual inputs and the textual descriptions contribute to improvements in trajectory prediction performance, and our qualitative analysis highlights how the model is able to exploit these additional inputs. 
Lastly, in this work we create and release the nuScenes-Text dataset, which augments the established nuScenes dataset with rich textual annotations for every scene, demonstrating the positive impact of utilizing VLM on trajectory prediction. Our project page is at \href{https://moonseokha.github.io/VisionTrap}{https://moonseokha.github.io/VisionTrap}.        

\keywords{Motion Forecasting \and Trajectory Prediction \and Autonomous Driving \and nuScenes-Text Dataset}

\vspace{-1.5em}
\end{abstract}

%% file: sec/1_introduction.tex
\section{Introduction}
\label{sec:introduction}
Predicting agents' future poses (or trajectories) is crucial for safe navigation in dense and complex urban environments. To achieve such task successfully, it is required to model the following aspects: (i) understanding individual's behavioral contexts (\eg, actions and intentions), (ii) agent-agent interactions, and (iii) agent-environment interactions (\eg, pedestrians on the crosswalk). Recent works~\cite{hivt, thomas, laformer, multipath, wayformer, qcnet, mmtransformer, HOME} have achieved remarkable progress, but their inputs are often limited -- they mainly use a high-definition (HD) map and agents' past trajectories from a detection and tracking system as inputs. 

HD map is inherently static, and only provide pre-defined information that limits their adaptability to changing environmental conditions like traffic near construction areas or weather conditions. They also cannot provide visual data for understanding agents' behavioral context, such as pedestrians' gazes, orientations, actions, gestures, and vehicle turn signals, all of which can significantly influence agents' behavior. Therefore, scenarios requiring visual context understanding may necessitate more than non-visual input for better and more reliable performance.
\begin{figure}[t]
    \centering
    \includegraphics[width=\linewidth]{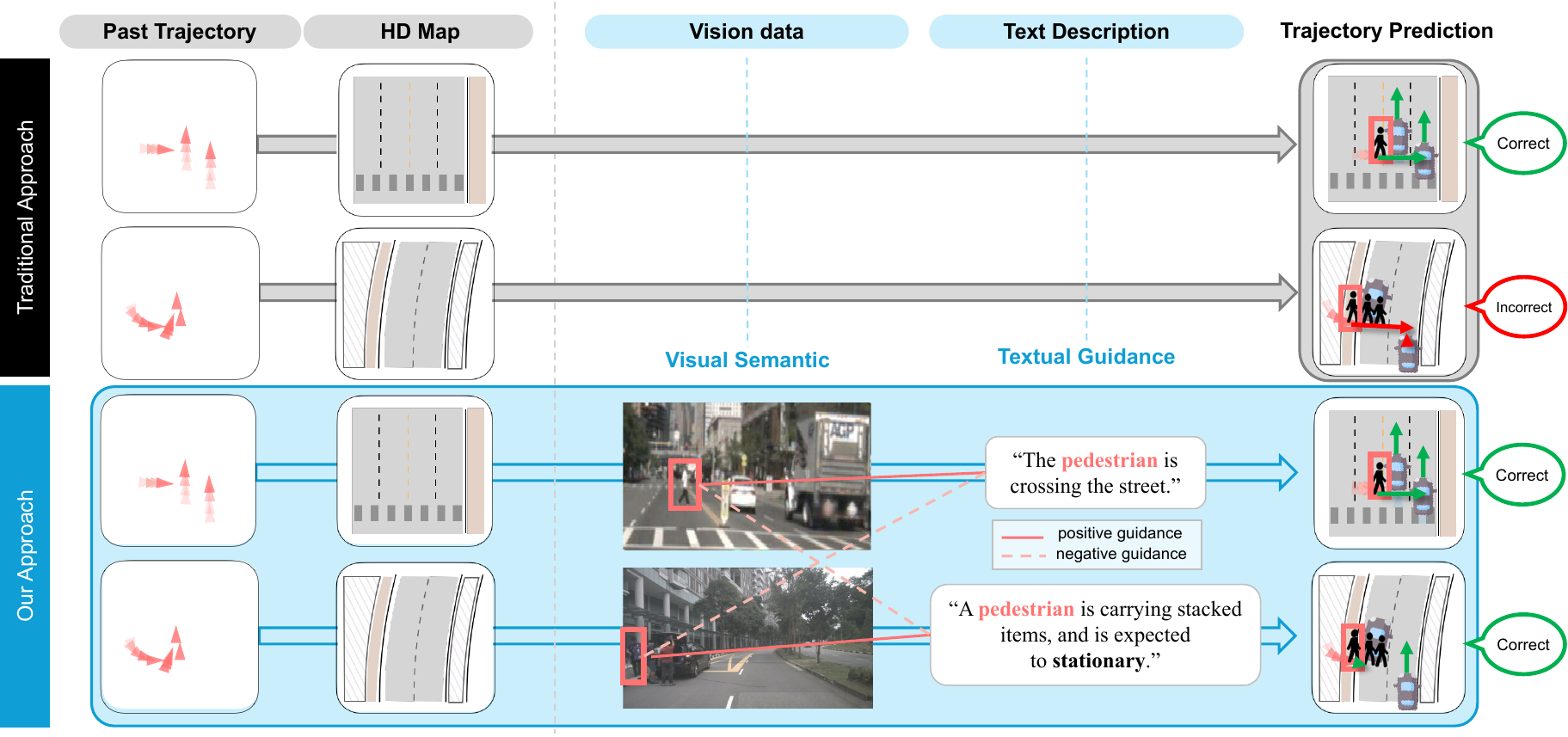}
    \vspace{-1.5em}
    \caption{Existing approaches are often conditioned only on agents' past trajectories and HD map to predict future trajectories. Here, we want to explore leveraging camera images and textual descriptions obtained from images to better learn the agent's behavioral context and agent-environment interactions by incorporating high-level semantic information into the prediction process, such as ``a pedestrian is carrying stacked items, and is expected to stationary.''}
    \label{fig:enter-label}
    \vspace{-2.2em}
\end{figure}

In this paper, we advocate for leveraging visual semantics in the trajectory prediction task. 
We argue that visual inputs can provide useful semantics, which non-visual inputs may not provide, for accurately predicting agents' future trajectories. 
Despite its potential advantages, only a few works~\cite{bifold, tpnet, pepscenes, titan, multimodal,spatiotemporal,pie} have used vision data to improve the performance of trajectory prediction in autonomous driving domain. 
Existing approaches often utilize images of the area where the agent is located or the entire image without explicit instructions on what information to extract. As a result, these methods tend to focus only on salient features, leading to sub-optimal performance. Additionally, because they typically rely solely on frontal-view images, it becomes challenging to fully recognize the surrounding driving environment.

To address these limitations and harness the potential of visual semantics, we propose \textbf{VisionTrap}, a vision-augmented trajectory prediction model that efficiently incorporates visual semantic information. To leverage visual semantics obtained from surround-view camera images, we first encode them into a composite Bird's Eye View (BEV) feature along with map data. Given this vision-aware BEV scene feature, we use a deformable attention mechanism to extract scene information from relevant areas (using predicted agents' future positions), and augment them into per-agent state embedding, producing scene-augmented state embedding. 
In addition, recent works~\cite{loki,intentnet,titan,spatiotemporal,pie} have shown that classifying intentions can improve model performance by helping predict agents' instantaneous movements. Learning with supervision of each agent's intention helps avoid training restrictions and oversimplified learning that may not yield optimal performance. 
However, annotating agents' intentions by dividing them into action categories involves inevitable ambiguity, which can be costly and hinder efficient scalability. Moreover, creating models that rely on these small sets can limit the model's expressiveness. Thus, as shown in Fig.~\ref{fig:enter-label}, we leverage textual guidance as supervision to guide the model in leveraging richer visual semantics by aligning visual features (\eg, an image of a pedestrian nearby a parked vehicle) with textual descriptions (\eg, ``a pedestrian is carrying stacked items, and is expected to stationary.''). 
While we use additional input data, real-time processing is crucial in autonomous driving. Therefore, we designed VisionTrap based on a real-time capable model proposed in this paper. VisionTrap efficiently utilizes visual semantic information and employs textual guidance only during training. This allows it to achieve performance comparable to high-accuracy, non-real-time single-agent prediction methods~\cite{p2t,mhajam} while maintaining real-time operation.

Since currently published autonomous driving datasets do not include textual descriptions, we created the nuScenes-Text dataset based on the large-scale nuScenes dataset~\cite{nuscenes}, which includes vision data and 3D coordinates of each agent. The nuScenes-Text dataset collects textual descriptions that encompass high-level semantic information, as shown in \cref{fig:sub_result}: ``A man wearing a blue shirt is talking to another man, expecting to cross the street when the signal changes.'' Automating this annotation process, we utilize both a Vision-Language Model (VLM) and a Large-Language Model (LLM).

Our extensive experiments on the nuScenes dataset show that our proposed text-guided image augmentation is effective in guiding our trajectory prediction model successfully to learn individuals' behavior and environmental contexts, producing a significant gain in trajectory prediction performance. 

%% file: sec/2_relatedwork.tex
\section{Related Work} \label{sec:related_work}

\vspace{-.5em}\myparagraph{Encoding Behavioral Contexts for Trajectory Prediction.}
Recent works in trajectory prediction utilize past trajectory observations and HD map to provide static environmental context. Traditional methods use rasterized Bird’s Eye View (BEV) maps with ConvNet blocks~\cite{multipath, HOME, covernet, trajectron++, air2}, while recent approaches employ vectorized maps with graph-based attention or convolution layers for better understanding complex topologies~\cite{vectornet, lanegcn, gohome, thomas, fjmp, trajectron++}. However, HD maps are static and cannot adapt to changes, like construction zones affecting agent behavior. To address this, some works~\cite{bifold, tpnet, titan, pepscenes, multimodal} aim to address these issues by utilizing images. To obtain meaningful visual semantic information about the situations an agent faces in a driving scene, it is necessary to utilize environmental information containing details from the objects themselves and from the environments they interact with. However, \cite{multimodal, bifold, titan} focus solely on extracting information about agents’ behavior using images near the agents, while \cite{tpnet, pepscenes} process the entire image at once and focus only on information about the scene without considering the parts that agents need to interact with. Therefore, in this paper, we propose an effective way to identify relevant parts of the image that each agent should focus on and efficiently learn semantic information from those parts.

\myparagraph{Scene-centric vs. Agent-centric.}
Two primary approaches to predicting road agents' future trajectories are scene-centric and agent-centric. Scene-centric methods~\cite{scenetransformer,narrowing,lanercnn} encode each agent within a shared scene coordinate system, ensuring rapid inference speed but may exhibit slightly lower performance than agent-centric methods. Agent-centric approaches~\cite{mmtransformer,poly,laformer,pgp,hivt} standardize environmental elements and separately predict agents' future trajectories, offering improved predictive accuracy. However, their inference time and memory requirements are linearly scaled with the number of agents in the scene, posing a scalability challenge in dense urban environments with hundreds of pedestrians and vehicles. Thus, in this paper, we focus on scene-centric approaches. 

\myparagraph{Multimodal Contrastive Learning.}
With the increasing diversity of data sources, multimodal learning has become popular as it aims to effectively integrate information from various modalities. One of the common and effective approaches for multimodal learning is to align the modalities in a joint embedding space, using contrastive learning~\cite{clip, align, mmclvis}.
Contrastive Learning (CL) pulls together the positive pairs and pushes away the negative pairs, constructing an embedding space that effectively accommodates the semantic relations among the representations. 
Although CL is renowned for its ability to create a robust embedding space, its typical training mechanism introduces sampling bias, unintentionally incorporating similar pairs as negative pairs~\cite{debiasedcl}. 
Debiasing strategies~\cite{debiasedcl,dclr,wcl,selecmix,debcse} have been introduced to mitigate such false-negatives, and it is particularly crucial in autonomous driving scenarios where multiple agents within a scene might have similar intentions in their behaviors.
In our work, we carefully design our contrastive loss by filtering out the negative samples that are considered to be false-negatives.
Inspired by~\cite{dclr, debcse}, we do this by utilizing the sentence representations and their similarities, and finally achieve debiased contrastive learning in multimodal setting.

%% file: sec/3_method.tex
\vspace{-1.2em}
\begin{figure*}[t]
    \centering
    \includegraphics[width=\linewidth]{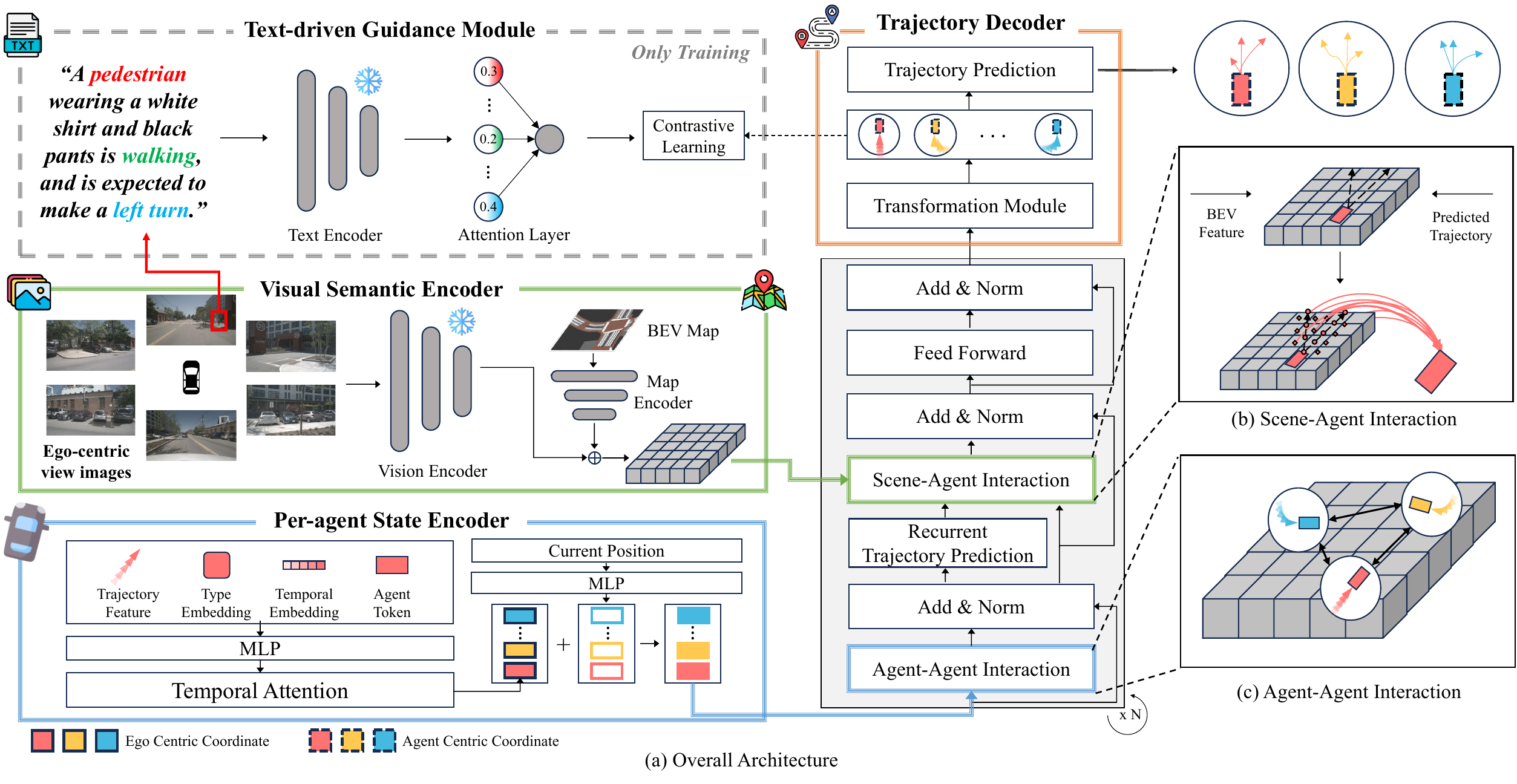}
    \vspace{-2.1em}
    \caption{An overview of VisionTrap, which consists of four main steps: (i) Per-agent State Embedding, which produces per-agent context features given agents' state observations; (ii) Visual Semantic Encoder, which transforms multi-view images with an HD map into a unified BEV feature, updating agents' state embedding via a deformable attention layer; (iii) Text-driven Guidance Module, which supervises the model to reason about detailed visual semantics and (iv) Trajectory Decoder, which predicts agents' the future poses in a fixed time horizon.}
    \label{fig:main}
    \vspace{-1.8em}
\end{figure*}
\section{Method}
This paper explores leveraging high-level visual semantics to improve the trajectory prediction quality. In addition to conventionally using agents' past trajectories and their types as inputs, we advocate for using visual data as an additional input to utilize agents' visual semantics. As shown in Fig.~\ref{fig:main}, our model consists of four main modules: (i) Per-agent State Encoder, (ii) Visual Semantic Encoder, (iii) Text-driven Guidance module, and (iv) Trajectory Decoder. Our {\em Per-agent State Encoder} takes as an input a sequence of state observations (which are often provided by a detection and tracking system), producing per-agent context features (Sec.~\ref{sec:traj_encoder}). In our {\em Visual Semantic Encoder}, we encode multi-view images (capturing the surrounding view around the ego vehicle) into a unified Bird's Eye View (BEV) feature, followed by concatenation with a dense feature map of road segments. Given this BEV feature, the per-agent state embedding is updated in the Scene-Agent Interaction module (Sec.~\ref{sec:vision_augmented_interaction}). We utilize {\em Text-driven Guidance module} to supervise the model to understand or reason about detailed visual semantics, producing richer semantics (Sec.~\ref{sec:text_guidance}). Lastly, given per-agent features with rich visual semantics, our {\em Trajectory Decoder} predicts the future positions for all agents in the scene in a fixed time horizon (Sec.~\ref{sec:traj_decoder}). 

\vspace{-1.2em}
\subsection{Per-agent State Encoder}\label{sec:traj_encoder}
\vspace{-.5em}
\myparagraph{Encoding Agent State Observations.}
Following recent trajectory prediction approaches~\cite{hivt, wayformer}, we first encode per-agent state observations (\eg, agent's observed trajectory and semantic attributes) provided by object detection and tracking systems. We utilize the geometric attributes with relative positions (instead of absolute positions) by representing the observed trajectory of agent $i$ as 
$\{p_i^t - p_i^{t-1}\}_{t=1}^{T}$ where $p_i^t=(x_i^t, y_i^t)$ is the location of agent $i$ 
in an ego-centric coordinate system at time step $t\in\{1, 2, \dots, T\}$. $T$ denotes the observation time horizon. Note that we use an ego-centric (scene-centric) coordinate system where a scene is centered and rotated around the current ego-agent's location and orientation. 
Given these geometric attributes and their semantic attributes $a_i$ (\ie, agent types, such as cars, pedestrians, and cyclists), per-agent state embedding $s_i^t\in\mathbb{R}^{d_s}$ for agent $i$ at time step $t$ is obtained as follows:
\begin{equation}
    \label{eq:eq1}
    s_{i}^{t} = f_{\text{geometric}}(p_i^t - p_i^{t-1}) + f_{\text{type}}(a_i) + f_{\text{PE}}(e^t),
\end{equation}
where $f_{\text{geometric}}: \mathbb{R}^{2}\rightarrow\mathbb{R}^{d_s}$, $f_{\text{type}}: \mathbb{R}^{1}\rightarrow\mathbb{R}^{d_s}$, and $f_{\text{PE}}: \mathbb{R}^{d_{pe}}\rightarrow\mathbb{R}^{d_{s}}$ are MLP blocks. Note that we use the learned positional embeddings $e^t\in\mathbb{R}^{d_{pe}}$, guiding the model to learn (and utilize) the temporal ordering of state embeddings. 

\myparagraph{Encoding Temporal Information.}
Following existing approaches~\cite{hivt, agentformer}, we utilize a temporal Transformer encoder to learn the agent's temporal information over the observation time horizon. Given the sequence of per-agent state embeddings $\{s_i^t\}_{t=1}^{T}$ and an additional learnable token $s^{T+1}\in\mathbb{R}^{d_s}$ stacked into the end of the sequence, we feed these input into the temporal (self-attention) attention block, producing per-agent spatio-temporal representations $s'_i\in\mathbb{R}^{d_s}$. 

\myparagraph{Encoding Interaction between Agents.}
We further use the cross-attention-based agent-agent interaction module to learn the relationship between agents. Further, as our model depends on the geometric attributes with relative positions, we add embeddings of the agents' current position $p^T_i$ to make the embeddings spatially aware, producing per-agent representation $z_i=s'_i + f_{loc}(p^T_i)$ where $f_{loc}: \mathbb{R}^{2}\rightarrow\mathbb{R}^{d_s}$ is another MLP block. This process is performed at once within the ego-centric coordinate system to eliminate the cost of recalculating correlation distances with other agents for each individual agent. The agent state embedding $z_i$ is used as the query vector, and those of its neighboring agents are converted to the key and the value vectors as follows:
\begin{equation}
    \label{eq:eq5}
    q^{\text{Interact}}_i = W^{\text{Interact}}_Q z_i, \quad k^{\text{Interact}}_{j} = W^{\text{Interact}}_K z_{j}, \quad v^{\text{Interact}}_{j} = W^{\text{Interact}}_V z_{j},
\end{equation}
where $W^{\text{Interact}}_Q, W^{\text{Interact}}_K, W^{\text{Interact}}_V\in\mathbb{R}^{d_{\text{Interact}}\times d_s}$ are learnable matrices.

\subsection{Visual Semantic Encoder}
\label{sec:vision_augmented_interaction}
\myparagraph{Vision-Augmented Scene Feature Generation.}
Given ego-centric multi-view images $\mathcal{I}=\{\mathcal{I}_j\}_{j=1}^{n_I}$, we feed them into Vision Encoder using the same architecture from BEVDepth~\cite{bevdepth}, to produce the BEV image feature as $B_I\in\mathbb{R}^{h\times w\times d_{\text{bev}}}$. Then, we incorporate the rasterized map information into the BEV embeddings to align $B_{I}$. We utilize CNN blocks with Feature Pyramid Network (FPN)~\cite{fpn} to produce another BEV feature $B_{\text{map}}\in\mathbb{R}^{h\times w\times d_{\text{map}}}$. Lastly, we concatenate all generated BEV features into a composite BEV scene feature $B=[B_I; B_{\text{map}}]\in\mathbb{R}^{h\times w\times (d_{\text{bev}}+d_{\text{map}})}$. In this process, we compute map aligned around the current location and direction of the ego vehicle only once, even in the presence of $n$ agents, as we adopt an ego-centric approach. This significantly reduces computational costs compared to agent-centric approaches, which require reconstructing and encoding map for each of the $n$ agents.

\myparagraph{Augmenting Visual Semantics into Agent State Embedding.}
When given the vision-aware BEV scene feature $B$, we use deformable cross-attention~\cite{deformdetr} module to augment map-aware visual scene semantics into the per-agent state embedding $z_i$, as illustrated in Fig. \ref{fig:main} (b). This allows for the augmentation of agent state embedding $z_i$.
Compared to commonly used ConvNet-based architectures \cite{multipath,HOME,covernet}, our approach leverages a wide receptive field and can selectively focus on scene feature, explicitly extracting multiple areas where each agent needs to focus and gather information. Additionally, as the agent state embedding is updated for each block, the focal points for the agent also require repeated refinement. To achieve this, we employ a Recurrent Trajectory Prediction module, which utilizes the same architecture as the main trajectory decoder(explained in \cref{sec:traj_decoder}). This module refines the agent's future trajectory $u^\text{aux}=\{u^\text{aux}_i\}_{i=1}^{T_f}$ by recurrently improving the predicted trajectories. These refined trajectories serve as reference points for agents to focus on in the Scene-Agent Interaction module, integrating surrounding information around the reference points into the agent's function. Our module is defined as follows:
\begin{equation}
    \label{eq:e2a}
    z_i^\text{scene} =  
    z_i^\text{interact} + \sum_{h=1}^{H}W_{h}\left [\sum_{o=1}^{O}\left(\alpha_{hio}{W}'_{h}\mathbf{B}_{\left (u^\text{aux}_i+\triangle u^\text{aux}_{hio}\right)}\right) \right],
\end{equation}
where $H$ denotes the number of attention heads and $O$ represents the number of offset points for every reference point $u^\text{aux}_{i}$ where we use an auxiliary trajectory predictor and use the agent's predicted future positions as reference points. Note that $W_h$ and $W'_h$ are learnable matrices, and $\alpha_{hio}$ is the attention weight for each learnable offset $\triangle u^\text{aux}_{hio}$ in each head. The number of attention points is typically set fewer than the number of surrounding road elements, reducing computational costs.

\begin{figure*}[t]
    \centering
    \includegraphics[width=.93\linewidth]{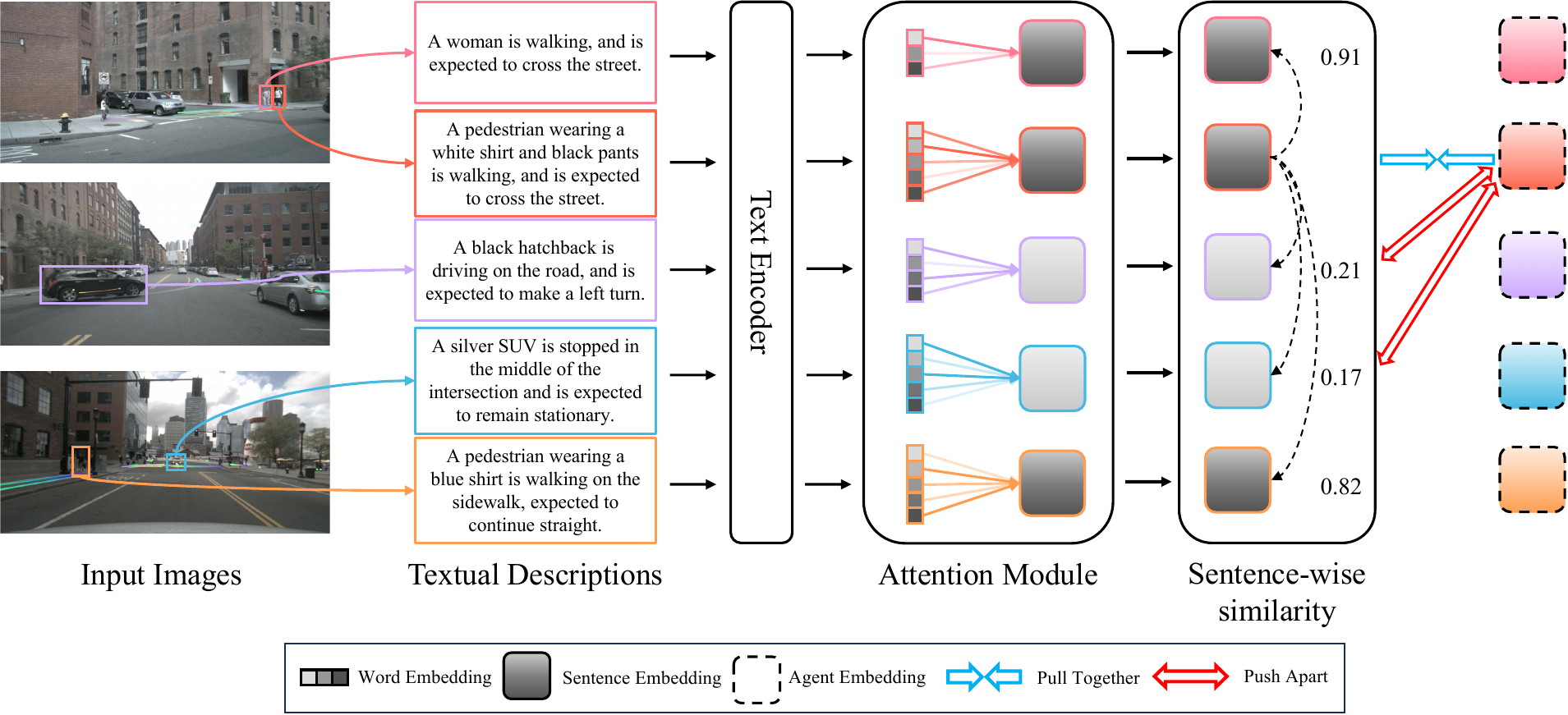}
    \vspace{-.8em}
    \caption{An overview of our Text-driven Guidance Module. We extract word-level embeddings using pretrained BERT~\cite{bert} as a text encoder, and then we use an attention module to aggregate these per-word embeddings into a composite sentence-level embedding. Based on the cosine similarity between these embeddings, we apply contrastive learning loss to ground textual descriptions into the agent's state embedding.
    }
    \label{fig:cl}
    \vspace{-1.5em}
\end{figure*}
\subsection{Text-driven Guidance Module}\label{sec:text_guidance}
\label{sec:contrast}
We observe that our visual semantic encoder simplifies visual reasoning about a scene to focus on salient visible features, resulting in sub-optimal performance in trajectory prediction. For instance, the model may primarily focus on the vehicle itself, disregarding other semantic details, such as ``a vehicle waiting in front of the intersection with turn signals on, expected to turn left.'' Therefore, we introduce the Text-driven Guidance Module to supervise the model, allowing the model to understand the context of the agents using detailed visual semantics. For this purpose, we employ multi-modal contrastive learning where positive pair is pulled together and negative pairs are pushed farther.
However, the textual descriptions for prediction tasks in the driving domain are diverse in expression, posing an ambiguity in forming negative pairs between descriptions. 

To address this, as shown in Fig.~\ref{fig:cl}, we extract word-level embeddings using BERT~\cite{bert}, and then we use a attention module to aggregate these per-word embeddings into a composite sentence-level embedding $\mathcal{T}_i$ for agent $i$. Given $\mathcal{T}_i$, we measure cosine similarity with other agents' sentence-level embeddings $\mathcal{T}_j$ for $j \neq i$, and we treat as negative pairs if $\text{sim}_\text{cos}\left(\mathcal{T}_i, \mathcal{T}_j\right) < \theta_\text{th}$ where $\theta_\text{th}$ is a threshold value (we set $\theta_\text{th}=0.8$ in our experiments). Further, we limit the number of negative pairs within a batch for stable optimization, which is particularly important as the number of agents in a given scene varies. Specifically, given an agent $i$, we choose top-$k$ sentence-level embeddings from $\{\mathcal{T}_j\}$ sorted in ascending order for $j \neq i$. Subsequently, we form a positive pair between the agent's state embedding $z_i^\text{scene}$ and corresponding textual embedding $\mathcal{T}_i$, while negative pairs as $z_i^\text{scene}$ and $\{\mathcal{T}_j\}_{j=1}^{k}$. Ultimately, we use the following InfoNCE loss~\cite{cpc} to guide agent's state embedding with textual descriptions:
\begin{equation}
    \label{eq:infonce}
    \mathcal{L}_\textnormal{cl}=-\log \frac{e^{\text{sim}_\text{cos}\left(z_i^\text{scene}, \mathcal{T}_i\right) / \tau}}{\sum_{j=1}^{k}e^{\text{sim}_\text{cos}\left(z_i^\text{scene}, \mathcal{T}_j\right) / \tau}},
\end{equation}
where $\tau$ is a temperature parameter used in the attention layer, enabling biasing the distribution of attention scores.

\subsection{Trajectory Decoder}
\label{sec:traj_decoder}

\begin{wrapfigure}{br}{.5\textwidth}
    \vspace{-3em}
    \centering
    \includegraphics[width=0.95\linewidth]{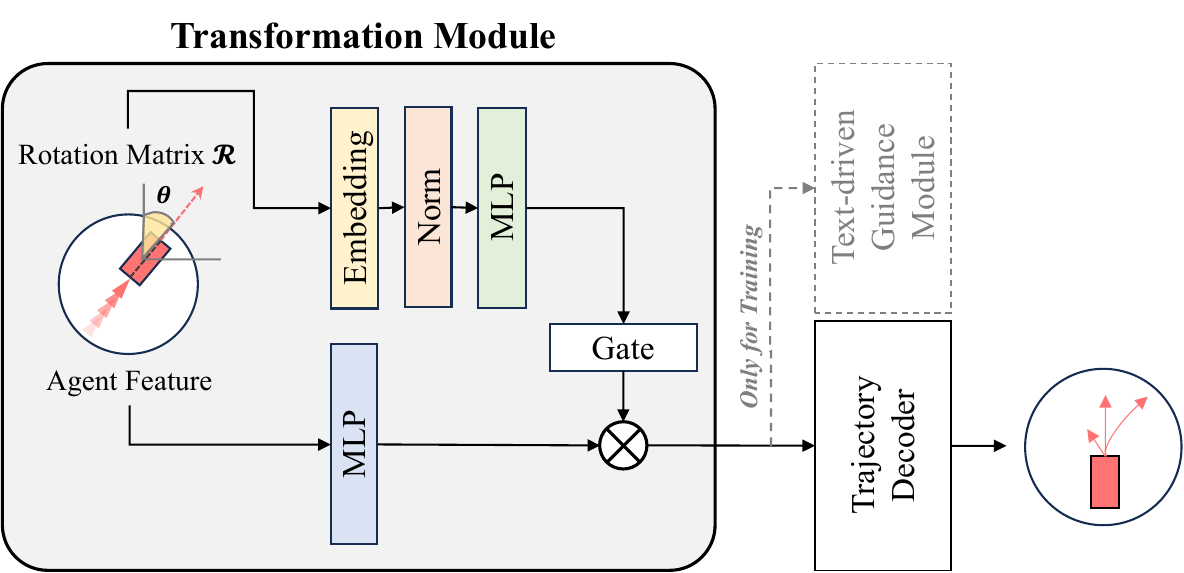}
    \vspace{-.5em}
    \caption{An overview of transformation module, which standardizes agents' orientation.}
    \label{fig:transformation}
    \vspace{-1.8em}
\end{wrapfigure} 
\subsubsection{Transformation Module.}
For fast inference speed and compatibility with ego-centric images, we adopt ego-centric approach in the State Encoder and Scene Semantic Interaction. However, as noted by Su~\etal~\cite{narrowing}, ego-centric approaches typically underperform compared to agent-centric approaches due to the need to learn invariance for transformations and rotations between scene elements. This implies that the features of agents with similar future movements are not standardized. 
Thus, prior to utilizing the Text-driven Guidance Module and predicting each agent's future trajectory, we employ the Transformation Module to standardize each agent's orientation, aiming to mitigate the complexity associated with learning rotation invariance.
This allows us to effectively apply the Text-driven Guidance Module, as we can make the features of agents in similar situations similar. 
As depicted in Fig.~\ref{fig:transformation}, the Transformation Module takes the agent's feature and rotation matrix $\mathcal{R}$ as input and propagates the rotation matrix to the agent's feature using a Multi-Layer Perceptron (MLP). This transformation enables the determination of which situations the agent's features face along the y-axis.

\myparagraph{Trajectory Decoder.}
Similar to \cite{multipath, multipath++, wayformer, covernet}, we use a parametric distribution over the agent's future trajectories $u=\{u_i\}_{i=1}^{T_f}$ for $u_i\in\mathbb{R}^{2}$ as Gaussian Mixture Model (GMM). We represent a mode at each time step $t$ as a 2D Gaussian distribution over a certain position with a mean $\mu_t\in\mathbb{R}^2$ and covariance $\Sigma_t\in\mathbb{R}^{2\times 2}$. Our decoder optimizes a weighted set of a possible future trajectory for the agent, producing full output distribution as
\begin{equation}
    p(u)= \sum_{m=1}^{M}\rho_m\prod_{t=1}^{T_f}\mathcal{N}(u_t-\mu_m^t, \Sigma_m^t),
\end{equation}
where our decoder produces a softmax probability $\rho$ over mixture components and Gaussian parameters $\mu$ and $\Sigma$ for $M$ modes and $T_f$ time steps.

\myparagraph{Loss Functions.}
We optimize trajectory predictions and their associated confidence levels by minimizing $\mathcal{L}_\textnormal{traj}$ to train our model in an end-to-end manner. We compute $\mathcal{L}_\textnormal{traj}$ by minimizing the negative log-likelihood function between actual and predicted trajectories and the corresponding confidence score, and it can be formulated as follows:
\begin{equation}
    \mathcal{L}_\textnormal{traj} = -\frac{1}{N} \sum_{i=1}^{N} \log \left( \sum_{m=1}^{M} \frac{\rho_{i,m}}{\sqrt{2b^2}} \exp \left( -\frac{(\mathbf{Y}_{i} - \hat{\mathbf{Y}}_{i,m})^2}2 \right) \right).
\end{equation}
Here, $b$ and $\mathbf{Y}$ represent the scale parameters and the real future trajectory, respectively. We denote predicted future positions as $\hat{\mathbf{Y}}_{i,m}$ and the corresponding confidence scores as $\rho_{i,m}$ for agent $i$ at future time step $t$ across different modes $m \in M$. Furthermore, we minimize an auxiliary loss function $\mathcal{L}_\textnormal{traj}^\text{aux}$ similar to $\mathcal{L}_\textnormal{traj}$ to train the trajectory decoder used by the Recurrent Trajectory Prediction module. Ultimately, our model is trained by minimizing the following loss $\mathcal{L}$, with $\lambda_\text{traj}^\text{aux}$ and $\lambda_\text{cl}$ controlling the strength of each loss term:
\begin{equation}
   \mathcal{L} = \mathcal{L}_\textnormal{traj} + \lambda_\text{traj}^\text{aux}\mathcal{L}_\textnormal{traj}^\text{aux} + \lambda_\text{cl}\mathcal{L}_\textnormal{cl}.
\end{equation}

%% file: sec/4_dataset.tex
\section{nuScenes-Text Dataset}\label{sec:dataset}
To our best knowledge, currently available driving datasets for prediction tasks lack textual descriptions of the actions of road users during various driving events.  While the DRAMA dataset~\cite{drama} offers textual descriptions for agents in driving scenes, it only provides a single caption for one agent in each scene alongside the corresponding bounding box. This setup suits detection and captioning tasks but not prediction tasks. To address this gap, we collect the textual descriptions for the nuScenes dataset~\cite{nuscenes}, which provides surround-view camera images, trajectories of road agents, and map data. With its diverse range of typical road agents activities, nuScenes is widely used in prediction tasks.

\begin{figure*}[t]
    \centering
    \includegraphics[width=\linewidth]{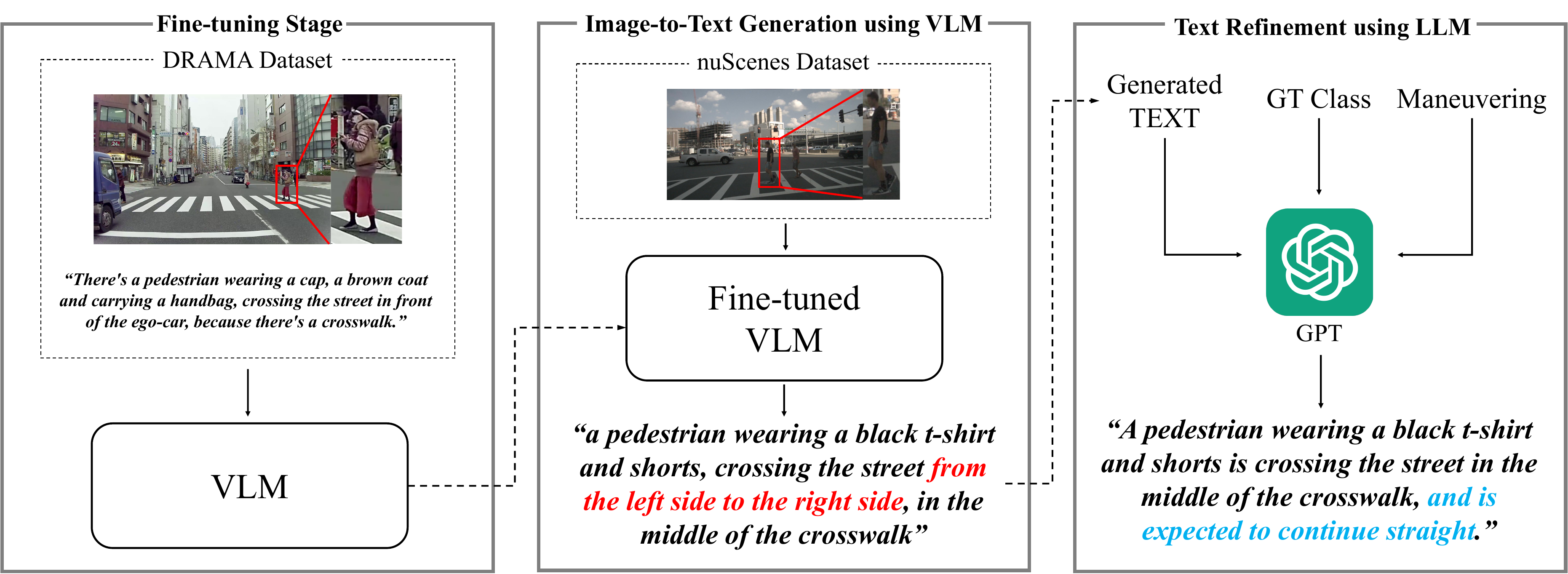}
    \vspace{-2em}
     \caption{To create the nuScenes-Text Dataset, three main steps are involved: (i) Fine-tuning stage using DRAMA Dataset~\cite{drama}, (ii) Image-to-Text Generation stage applying the fine-tuned VLM to the nuScenes Dataset~\cite{nuscenes}, and (iii) Text Refinement process using ground truth information (\eg GT class, Maneuvering) along with generated text and GPT~\cite{gpt3}. The \textcolor{red}{red} color indicates that needs to be filtered out, while the \textcolor{cyan}{cyan} color indicates additional content related to the intention.}
    \label{fig:text_generation}
    \vspace{-2.2em}
\end{figure*}
\myparagraph{Textual Description Generation.}
We employ a three-step process for generating textual descriptions of agents from images, as illustrated in Fig.~\ref{fig:text_generation}. Initially, we employ a pre-trained Vision-Language Model (VLM) BLIP-2~\cite{blip2}. However, it often underperforms in driving-related image-to-text tasks. To address this, we fine-tune the VLM with the DRAMA dataset~\cite{drama}, containing textual descriptions of agents in driving scenes. We isolate the bounding box region representing the agent of interest, concatenate it with the original image (Fig.~\ref{fig:text_generation}), and leverage the fine-tuned VLM to generate descriptions for each agent separately in the nuScenes dataset~\cite{nuscenes} as an image captioning task. However, the generated descriptions often lack correct action-related details, providing unnecessary information for prediction. 
To address shortcomings, we refine generated texts using GPT~\cite{gpt3}, a well-known Large Language Model (LLM).
Inputs include the generated text, agent type, and maneuvering. Rule-based logic determines the agent's maneuvering (\eg, stationary, lane change, turn right). We use prompts to correct inappropriate descriptions, aiming to generate texts that provide prediction-related information on agent type, actions, and rationale. Examples are provided in Fig.~\ref{fig:dataset}, with additional details (\eg, rule-based logic, GPT prompt) in the supplemental material.

\begin{figure*}[t]
    \centering
    \includegraphics[width=\linewidth]{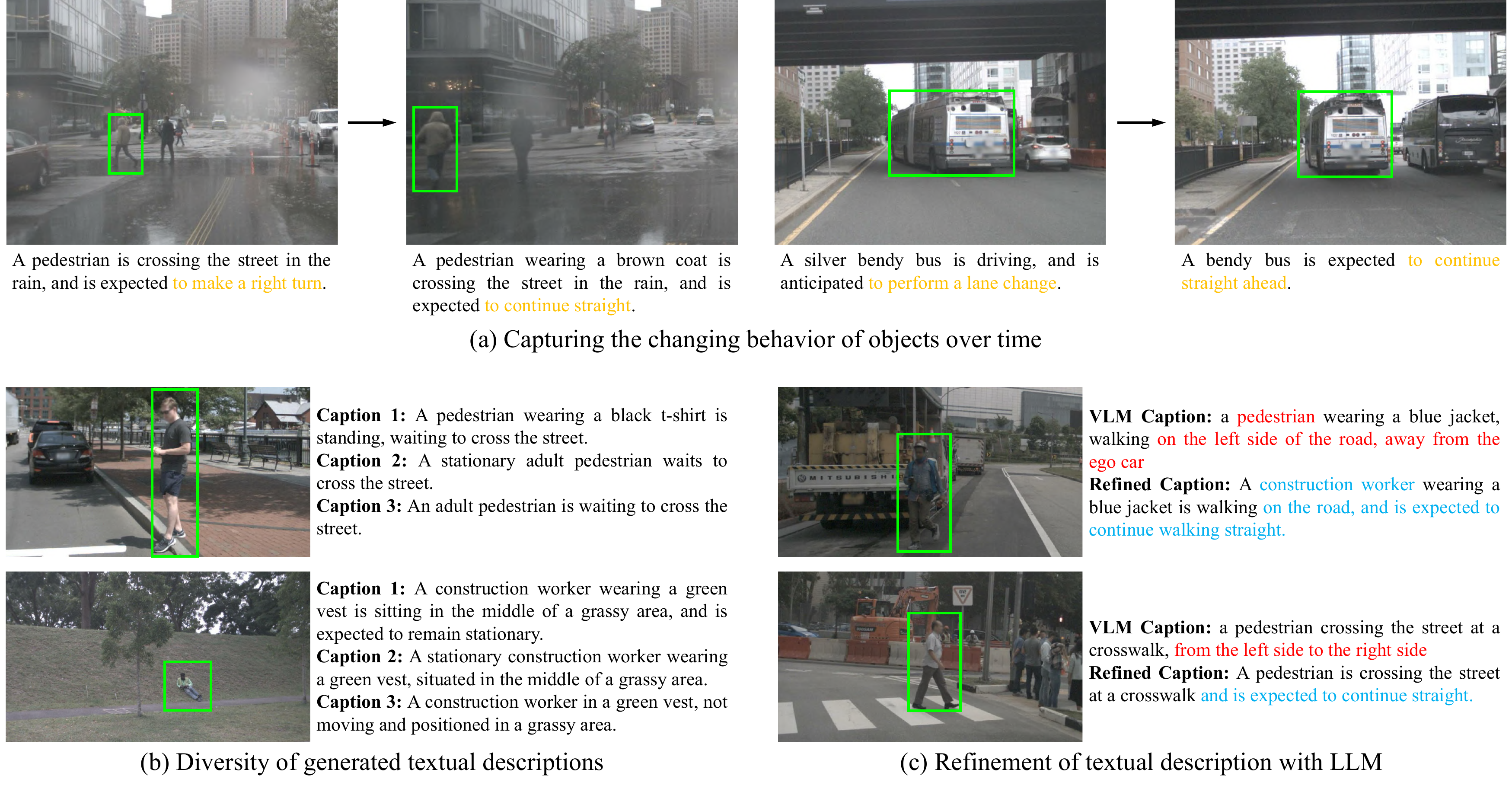}
    \vspace{-2em}
    \caption{Examples of our generated textual descriptions}
    \label{fig:dataset}
    \vspace{-2.1em}
\end{figure*}

\myparagraph{Coverage of nuScenes-Text Dataset.}
In this section, we demonstrate how well our created nuScene-Text Dataset encapsulates the context of the agent, as depicted in Fig.~\ref{fig:dataset}, and discuss the coverage and benefits of this dataset. Fig.~\ref{fig:dataset}a represents the contextual information of the agent changing over time in text form.
This attribute assists in accurately predicting object trajectories under behavioral context changes. We also demonstrate in \cref{fig:dataset}b that distinctive characteristics of each object can be captured (\eg, ``A pedestrian waiting to cross the street.'', ``A construction worker sitting on the lawn.'') and generate three unique textual descriptions for each object, showcasing diverse perspectives. 
Additionally, to enhance text descriptions when the VLM generates incorrect agent types, behavior predictions, or harmful information, such as ``from the left side to the right side'', which can be misleading due to the directional variation in BEV depending on the camera's orientation, we refine the text using an LLM. This refinement process aims to improve text quality for identifying driving scenes through surround images. \cref{fig:dataset}c illustrates this improvement process, ensuring the relevance and accuracy of text by removing irrelevant details (indicated by \textcolor{red}{red}) and adding pertinent information (indicated by \textcolor{cyan}{cyan}).
\begin{wrapfigure}{r}{.4\textwidth}
    \vspace{-2.em}
    \centering
    \includegraphics[width=0.98\linewidth]{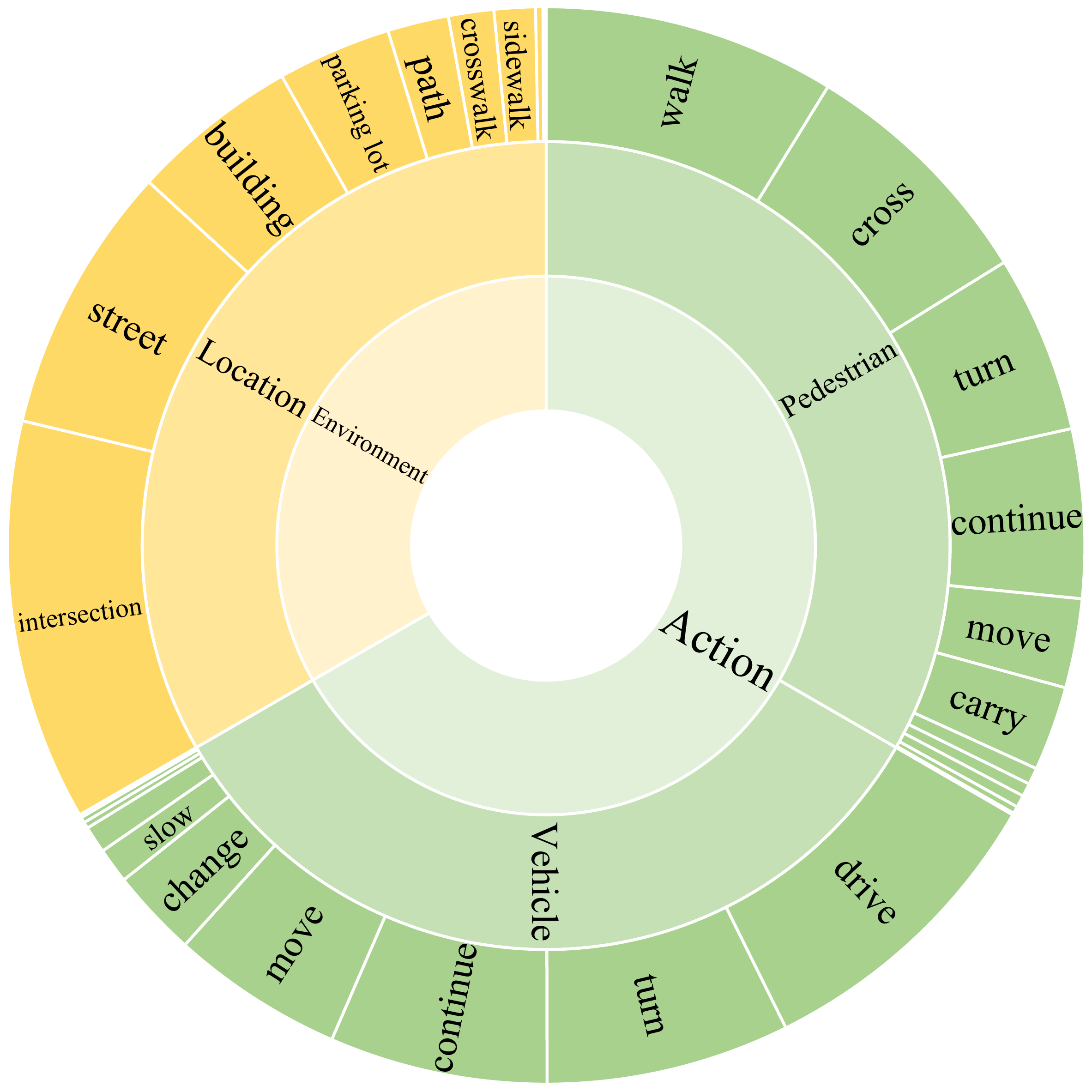}
    \vspace{-1.2em}
    \caption{Frequency of words}
    \label{fig:graph_top}
    \vspace{-2em}
\end{wrapfigure} 

\myparagraph{Dataset Statistics.}
Our created dataset contains 1,216,206 textual descriptions for 391,732 objects (three for each object), averaging 13 words per description. In Fig.~\ref{fig:graph_top}, we visualize frequently used words, highlighting the dataset's rich vocabulary and diversity. Further, we conduct a human evaluation using Amazon Mechanical Turk (Mturk) to quantitatively evaluate image-text alignments. 5 human evaluators are recruited, and it is performed on a subset of 1,000 randomly selected samples. Each evaluator is presented with the full image, cropped object image, and corresponding text and asked the question: ``Is the image well-aligned with the text, considering the reference image?''.
The results show that 94.8\% of the respondents chose `yes', indicating a high level of accuracy in aligning images with texts. All results are aggregated through a majority vote. Further details on the nuScenes-Text Dataset are provided in the supplemental material.

%% file: sec/5_experiment.tex
\section{Experiments}
\label{sec:experiment}
\vspace{-.5em}
\myparagraph{Dataset.}
We conduct experiments using the nuScenes dataset~\cite{nuscenes}, which offers two versions: (i) a dataset dedicated to a trajectory prediction task and (ii) a whole dataset. While the former focuses solely on single-agent prediction tasks, the latter is more suitable for our purposes. Therefore, we provide scores for both datasets in our experiments. Further implementation, evaluation, and dataset details can be found in the supplemental material.

\begin{figure}[t]
    \includegraphics[width=\linewidth]{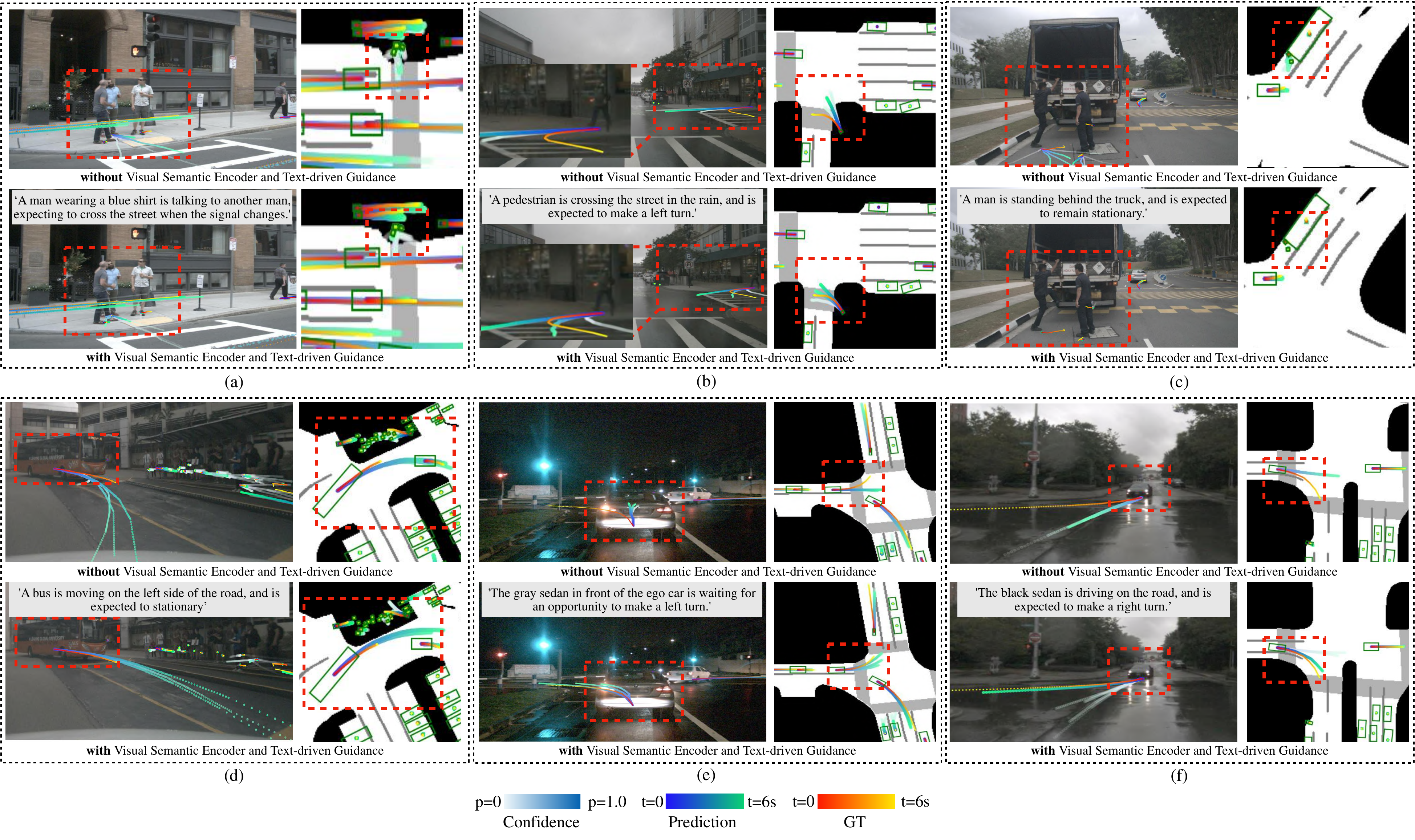}
    \vspace{-2em}
    \caption{Examples of trajectory prediction outputs in six different scenarios. 
    The examples on the top row represent scenarios with pedestrians, while those on the bottom row have vehicles. We also provide ground truth textual descriptions about an object in a red box, which were not seen during inference.}
    \label{fig:sub_result}
    \vspace{-1.8em}
\end{figure}

\myparagraph{Qualitative Analysis.}
Fig.~\ref{fig:sub_result} presents the results of VisionTrap on nuScenes dataset~\cite{nuscenes}, demonstrating the impact of Visual Semantic Encoder and Text-driven Guidance Module on agent trajectory prediction.

The top row shows improved results of pedestrians. For (a), while the result without visual information predicts the man will cross the crosswalk, the prediction with visual information indicates the man will remain stationary due to red traffic light and people talking to each other rather than trying to cross the road. 
(b) presents how gaze and body orientation help in predicting the pedestrian's intention to walk towards the crosswalk, and (c) provides visual context of the man getting on a stationary vehicle, implying the trajectory of the man would remain stationary as well.
The following row exhibits the improved prediction results of vehicles. In (d), understanding that the people are standing at a bus stop enables the model to make a reasonable prediction for the bus. (e) gives a visual cue of turn signal, indicating the vehicle's intention of turning left. Lastly, visual context in (f) leads to a more stable prediction of the vehicle turning right, as the image clearly shows the vehicle is directed towards its right.

These examples highlight the crucial role of visual data in improving trajectory prediction accuracy, offering insights that cannot obtained from non-visual data. Further qualitative analysis details are available in the supplemental material.

{
\setlength{\tabcolsep}{12pt}
\begin{table*}[t]    
    \centering
    \caption{Trajectory prediction performance comparison on nuScenes~\cite{nuscenes} dataset regarding ADE$_{10}$, MR$_{10}$, and FDE$_{1}$. Inference times are reported in milliseconds (msec), measured based on 12 agents using a single RTX 3090 Ti GPU.}
    \vspace{-.5em}
    \renewcommand{\arraystretch}{1.2}
        \addtolength{\tabcolsep}{-4pt}
    \resizebox{\linewidth}{!}{%
        \begin{tabular}{lcc|ccccccc}
        \toprule
        Model & \begin{tabular}[c]{@{}c@{}}Prediction\\ Method\end{tabular}& \begin{tabular}[c]{@{}c@{}}Using\\ Map Data\end{tabular} & \begin{tabular}[c]{@{}c@{}}Time $\downarrow$ \\ (msec)\end{tabular}  & ADE$_{10}$ $\downarrow$ & MR$_{10}$ $\downarrow$ & FDE$_{1}$ $\downarrow$  & \\ \midrule\midrule 
        Multipath~\cite{multipath}    & single & \ding{52} & 87 &  1.50 & 0.74 & -\\  
        MHA-JAM~\cite{mhajam}    & single & \ding{52} & - & 1.24 & 0.46 &  8.57  \\ 
        P2T~\cite{p2t}     & single& \ding{52} &  116  & 1.16 & 0.46 & 10.5  \\  
        PGP~\cite{pgp}      & single & \ding{52} &  215 &  1.00 & 0.37 &  7.17   \\ 
        LAformer~\cite{laformer}   & single   & \ding{52}  &  115 &  0.93& 0.33 &  - \\\midrule  
        \rowcolor[gray]{0.95} Trajectron++~\cite{trajectron++}   & multi & \ding{52} & 38 & 1.51 & 0.57 &  9.52 & \\  
        \rowcolor[gray]{0.95} AgentFormer~\cite{agentformer}     & multi & \ding{52}  &  107 & 1.45 & - &  -  &Average \\
        \rowcolor[gray]{0.95} VisionTrap baseline &  multi &  & 13  & 1.48  & 0.56 & 10.75 & improvement:\\ \midrule  
        \rowcolor[gray]{0.95}  + Map Encoder  &  multi & \ding{52} & 21 & 1.40 & 0.53 & 10.41 & 4.65\% \\  
        \rowcolor[gray]{0.95}  + Visual Semantic Encoder &  multi & \ding{52} & 53 & 1.23  & 0.36 & 9.32 & 21.97\% \\  
        \rowcolor[gray]{0.95}  + Text-driven Guidance (Ours)  &  multi & \ding{52} & 53 & 1.17  & 0.32 & 8.72 & 27.56\% \\  
        \bottomrule
    \end{tabular}}
    \label{table:comparison_with_sota}
    \vspace{-1.8em}
\end{table*}
}
\myparagraph{Quantitative Analysis.}
Tab.~\ref{table:comparison_with_sota} compares our model with other methods for single and multi-agent prediction. Our query-based prediction model designed to effectively utilize visual semantic information and Text-driven Guidance Module, which we use as baseline, achieves the fastest inference speed.
We also demonstrate that the Visual Semantic Encoder significantly improves performance, especially when combined with the Text-driven Guidance Module, yielding comparable results to existing single-agent prediction methods with better miss rate performance, while still maintaining real-time operation. 
These results suggest that vision data provides additional information inaccessible to non-vision data, and textual descriptions derived from vision data effectively guide the model.

\begin{wraptable}{r}{0.5\textwidth}  
    \vspace{-3.em}
    \caption{Ablation study of variant models on nuScenes~\cite{nuscenes} whole dataset.}
    \centering
    \resizebox{\linewidth}{!}{%
    \begin{tabular}{lccc}
    \toprule
    Method & ADE$_{10}$ $\downarrow$ & FDE$_{10}$ $\downarrow$ & MR$_{10}$ $\downarrow$ \\\midrule
    VisionTrap baseline & 0.425 & 0.641 & 0.081 \\
    + Map Encoder & 0.407 & 0.601 & 0.075 \\
    + Visual Semantic Encoder & 0.382 & 0.551 & 0.056 \\
    + Text-driven Guidance (Ours) & \textbf{0.368} & \textbf{0.535} & \textbf{0.051} \\ 
    \bottomrule
    \end{tabular}}
    \vspace{-2.5em}
    \label{tab:ablation}
\end{wraptable} 

Since our method employs egocentric surround-view images, it is feasible to effectively predict for all observed agents in the scene.
We utilize the nuScenes dataset covering all scenes, enabling comprehensive evaluation of all observed agents (refer to Tab. \ref{tab:ablation}). This demonstrates the contributions of all proposed components to predicting all agents in the scene.

Finally, we emphasize that the purpose of this study is not to achieve state-of-the-art performance. Instead, our aim is to demonstrate that vision information, often overlooked in trajectory prediction tasks, can provide additional insights. These insights are inaccessible from non-vision data, thereby enhancing performance in trajectory prediction tasks.
This is our original motivation for this task, and the results in \cref{fig:sub_result}, \cref{table:comparison_with_sota} and \cref{tab:ablation} provide justification for our method.

\begin{figure}[t]
    \includegraphics[width=\linewidth]{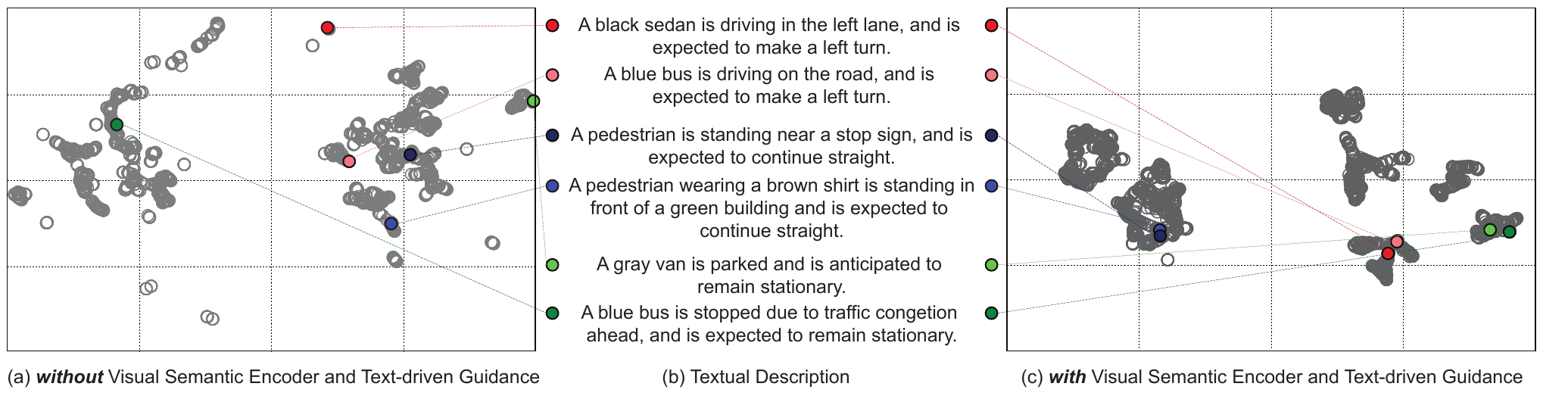}
    \vspace{-1.8em}
    \caption{UMAP~\cite{umap} visualizations for per-agent state embeddings from models (a) without and (c) with leveraging visual and textual semantics. (b) We also provide corresponding ground truth textual descriptions.}
    \label{fig:const_visualization}
    \vspace{-1.5em}
\end{figure}

\myparagraph{UMAP Visualization.}
We observe an overall improvement in clustering of agent state embeddings when leveraging visual and textual semantics in \cref{fig:const_visualization}. Furthermore, extracting textual descriptions of agents within the same cluster group is shown to exhibit similar situations. This indicates that state embeddings for agents in similar situations are located in a similar embedding space.

\begin{wraptable}{r}{0.5\textwidth}  
    \vspace{-3.5em}
    \caption{Performance comparison to analyze the effect of each component of Text-Based Guidance Module on the nuScenes~\cite{nuscenes} all dataset.} 
    \centering
    \resizebox{\linewidth}{!}{%
    \begin{tabular}{lccc}
    \toprule
    Method & ADE$_{6}$ $\downarrow$ & FDE$_{6}$ $\downarrow$ & MR$_{6}$ $\downarrow$ \\\midrule
    A. CLIP loss & 0.51 & 0.79 & 0.10 \\
    B. Ours w/ symmetric loss & 0.50 & 0.76 & 0.10 \\
    C. Ours w/o refining negative pair & 0.49 & 0.72 & 0.09 \\
    D. Ours w/o top-k algorithm & 0.46 & 0.67 & 0.08 \\ \midrule
    E. Ours & \textbf{0.44} & \textbf{0.66} & \textbf{0.07} \\ 
    \bottomrule
    \end{tabular}}
    \vspace{-2em}
    \label{tab:ablation_constras}
\end{wraptable}
\myparagraph{Analyzing the Text-driven Guidance Module.}
To analyze the effect of each component of the proposed Text-Based Guidance Module, we removed each factor to see how the model performs, as shown in Tab.~\ref{tab:ablation_constras}.
In the case of A, we use simple symmetric contrastive loss that is used in~\cite{clip}. 
However, our loss adopts asymmetric form of contrastive loss that only calculates softmax probabilities in one direction. 
B gives the result of incorporating symmetric loss in our loss design.
C shows the result of removing the stage of negative pair refinement, allowing potential false-negatives.
In D, we skip the process of ascending sorting and limiting the number of negative pairs. Removing these steps causes variance in number of agents considered each scene, leading to different scales of loss. 
In the end, our asymmetric contrastive loss with negative pairs refined and its number constrained demonstrated the best performance across all metrics.

%% file: sec/6_conclusion.tex
\section{Conclusion}
\vspace{-.2em}
\label{sec:conclusion}
In this paper, we introduced an novel approach called VisionTrap to trajectory prediction by incorporating visual input from surround-view cameras. This enables the model to leverage visual semantic cues, which were previously inaccessible to traditional trajectory prediction methods. Additionally, we utilize text descriptions produced by a VLM and refined by a LLM to provide supervision, guiding the model in learning from the input data. Our thorough experiments demonstrate that both visual inputs and textual descriptions contribute to enhancing trajectory prediction performance. Furthermore, our qualitative analysis shows how the model effectively utilizes these additional inputs.

%% file: src_supple/sec/setup.tex
\myparagraph{Dataset.}
Our proposed approach is developed and evaluated utilizing the widely employed nuScenes~\cite{nuscenes} dataset, which encompasses 1000 diverse scenes from Boston and Singapore. Annotations cover 10 classes for object detection, including car, truck, bus, trailer, construction vehicle, pedestrian, motorcycle, bicycle, barrier, and traffic cone. It also provides ego-centric surround-view images and HD map. In nuScenes, the model is trained with a 2-second history to predict a 6-second future trajectory. Unlike existing works~\cite{laformer,pgp,frm,thomas} that report about single-agent prediction performance, our research takes a different approach. Instead of utilizing only the dataset provided for the prediction task, we used the entire nuScenes dataset for training to conduct a multi-agent prediction approach that considers all agents in a scene simultaneously. Therefore, our nuScenes-Text dataset used for this study is created to cover all scenes in the nuScenes dataset. The Vision Language Model BLIP-2~\cite{blip2} (VLM) used to generate this text is trained on the DRAMA~\cite{drama} dataset, which provides an image of the driving environment, bounding box pointing to specific agent, and text representing this agent.
To accurately use textual descriptions obtained from fine-tuned VLM, we refine the descriptions using GPT~\cite{gpt3}. We also present metrics for all agents and metrics specifically for agents involved in the prediction task, offering a comprehensive evaluation.

\myparagraph{Evaluation Metrics.}
Our model is evaluated using standard metrics for trajectory prediction, including minimum Average Displacement Error (ADE), minimum Final Displacement Error (FDE), and Miss Rate (MR). These metrics quantify the average and final displacement errors between the true trajectory and the best prediction sample. MR further denotes the percentage of scenarios where the distance between the endpoint of the true trajectory and the best prediction exceeds a 2m threshold. 

\begin{equation}
    ADE = \frac{1}{T}\sum_{t=T_{curr}+1}^{T_{Fin}}\left\|\hat{Y}_{(k)}^{t} - Y^{t} \right\|_{2}
\end{equation}
\begin{equation}
    FDE = \left\|\hat{Y}_{(k)}^{T_{Fin}} - Y^{T_{Fin}} \right\|_{2}
\end{equation}
Here, $\hat{Y}_{(k)}^{t}$ denotes the predicted position of the agent at timestep $t$ in the $(k)$-th mode, and $Y^{t}$ represents the ground truth position at timestep $t$. The $(k)$ represents the mode with the smallest error when compared to the ground truth, while $T$ indicates the number of timesteps to be predicted.
Additionally, $T_{Fin}$ represents the timestep at which the prediction concludes, while $T_{curr}$ indicates the current timestep.

\myparagraph{Implementation Details.}
We train the model for 48 epochs using AdamW optimizer~\cite{adamw} and four RTX 3090 Ti GPUs. The model has 32 batch sizes, $5 \times 10^{-4}$ initial learning rates, $1\times 10^{-4}$ weight decay, and 0.1 dropout rates. To manage the learning rate, we adopt the cosine annealing scheduler~\cite{cosschedule}. For consistency, we set the number of offsets for deformable attention in the Scene-Agent Interaction Module, denoted as $O$, to 4. Additionally, augmentation techniques, including rotation within (-22.5, 22.5) degrees and excluding a random agents (10\% of all agents in scene) from the loss calculation, are used to prevent overfitting and increase the generalization performance of the model.

%% file: src_supple/sec/dataset.tex
\begin{figure*}[t]
    \centering
    \includegraphics[width=\linewidth]{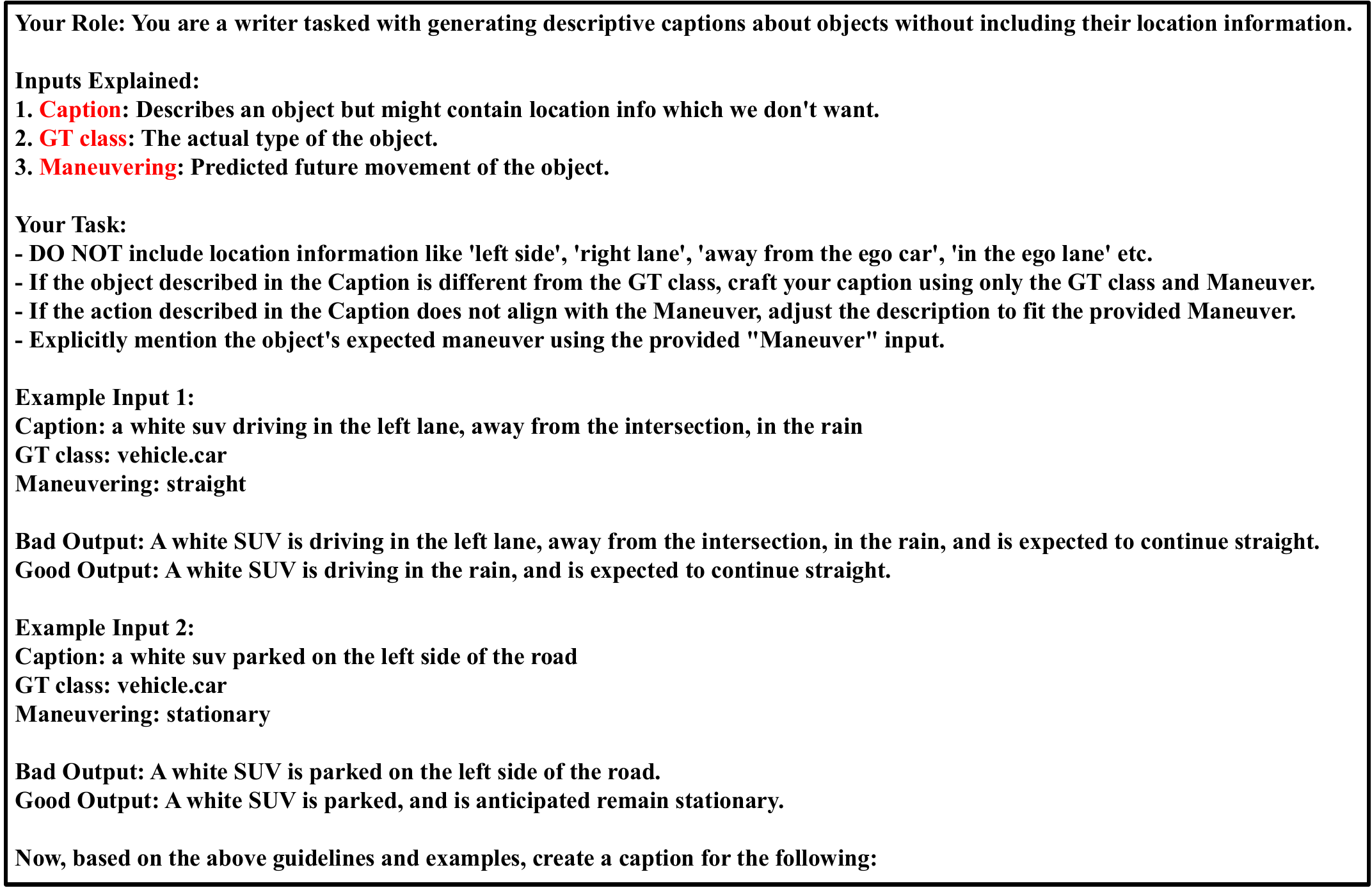}
    \vspace{-2em}
    \caption{Example of prompt given to the LLM, specifically designed to generate accurate descriptions. Inputs for this prompt, highlighted in red for emphasis, include the caption obtained from VLM, the object's class (GT class), and maneuvering information.}
    \label{fig:supple_prompt}
    \vspace{-1.5em}
\end{figure*}
\begin{figure*}[t]
    \centering
    \includegraphics[width=\linewidth]{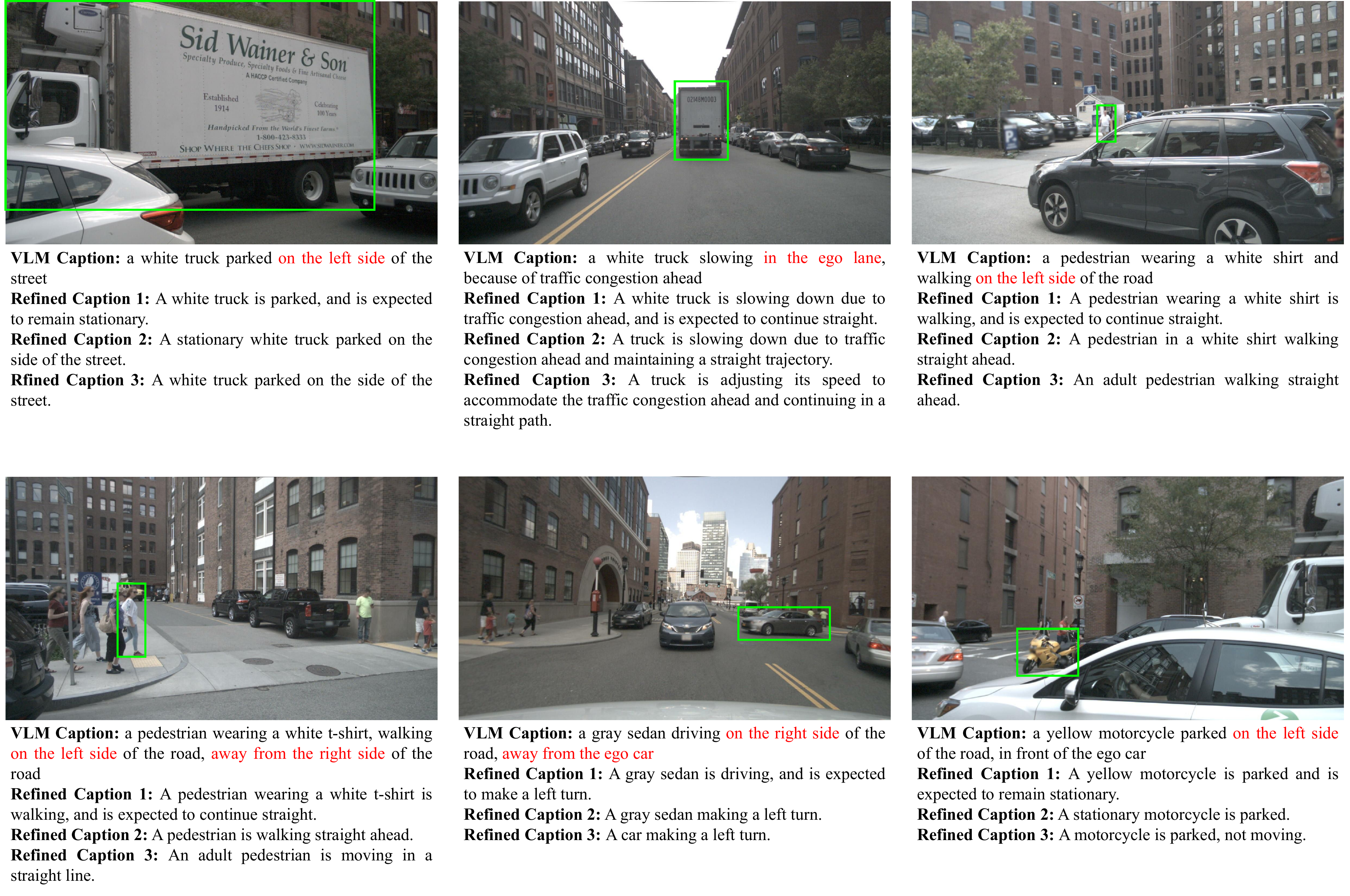}
    \vspace{-2em}
    \caption{Example of captions for objects in the nuScenes~\cite{nuscenes} dataset are provided in ego-centric surround-view images from a single scene. These captions describe each agent within the images, and each agent is accompanied by three versions of text.}
    \label{fig:supple_dataset}
    \vspace{-1.5em}
\end{figure*}

\myparagraph{Prompt Engineering for LLM.}
We utilize the Large Language Model (LLM) GPT to refine textual descriptions obtained from VLM regarding issues stemming from the domain gap between datasets or completely missing parts, as well as inaccurate location information caused by the characteristics of surround view images (see \cref{fig:supple_dataset}). To enhance the quality of pseudo-text, we meticulously design prompt for LLM such as \cref{fig:supple_prompt}. 
The primary challenges in this improvement process involved i) removing inaccurate location information such as `left,' `right lane,' or `ego lane,' caused by the characteristics of surround view images (see \cref{fig:supple_dataset}) and ii)  refining parts that are incorrectly predicted or completely missing due to domain gaps between datasets. Given that the nuScene dataset includes not only a front view but also a surround view, including back view, i) is crucial to avoid confusion in the model caused by these location details. Additionally, for ii), we explicitly integrate task details such as maneuvering and agent types to eliminate hallucinations and generate clear information. Finally, we include examples of both effective (`good') and ineffective (`bad') outputs to optimize the capabilities of the LLM.
\begin{figure*}[t]
    \centering
    \includegraphics[width=\linewidth]{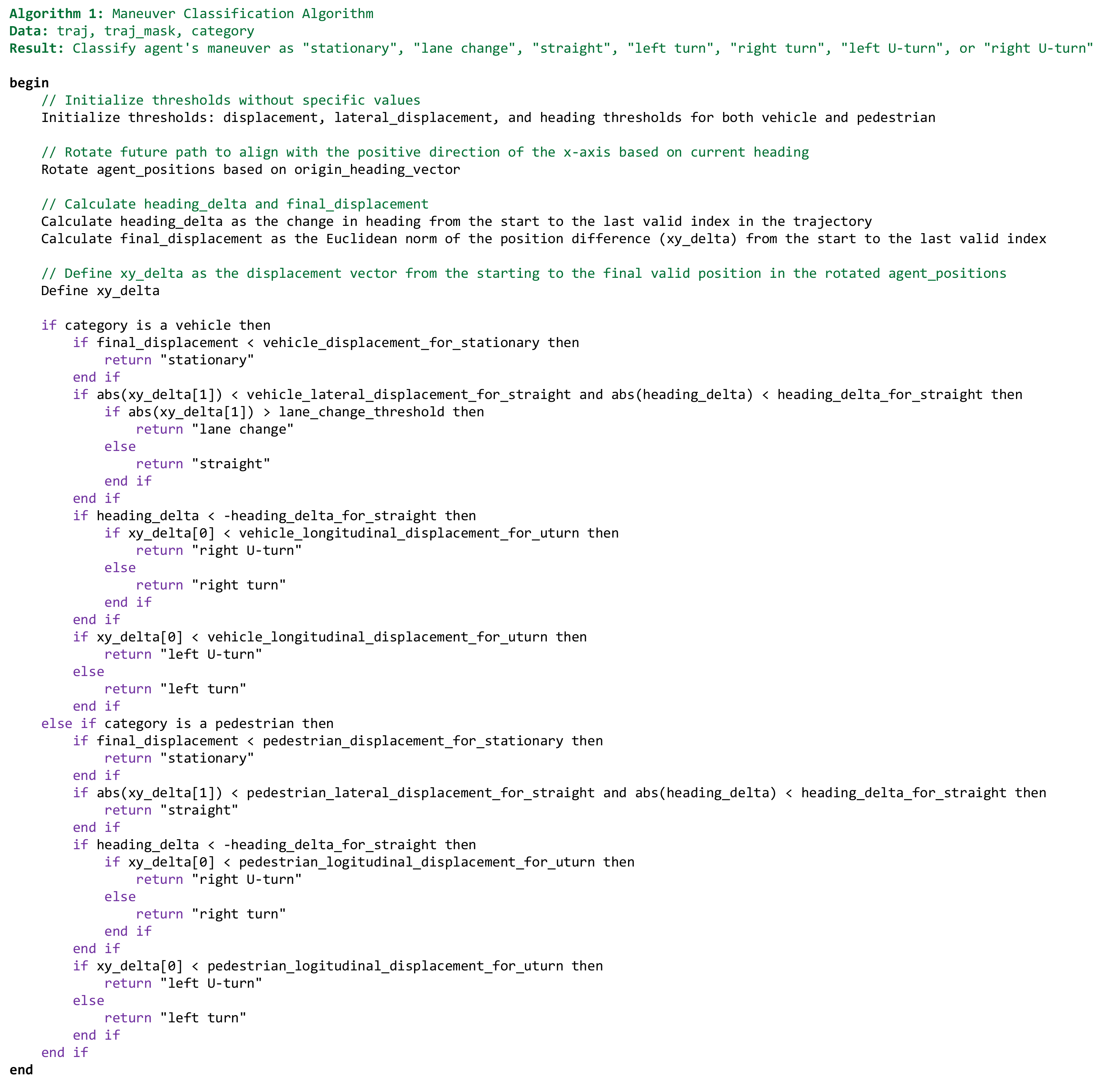}
    \caption{Maneuver Classification Algorithm. This algorithm shows the process for classifying the maneuvers of agents based on their future path and heading vector.}
    \label{fig:supple_algo}
    \vspace{-1.5em}
\end{figure*}
\myparagraph{Maneuvering extraction Algorithm.}
To integrate information about the intention of each agent into our generated text dataset, we utilize the maneuvering attribute. We classify the maneuvering of the agent based on the actual future trajectory. Maneuvering is defined by comparing the initial position and orientation with the final position and orientation. The generated maneuvering information is provided to the LLM to offer insights into the agent's intention. Therefore, the refined text, including information on the agent's characteristic points, current movement, and future intention, may be utilized, thereby contributing to enhancing the performance of the model. The maneuvering extraction algorithm can be observed in \cref{fig:supple_algo}.

\clearpage

\begin{figure*}[t]
    \vspace{1.3em}
    \centering
    \includegraphics[width=\linewidth]{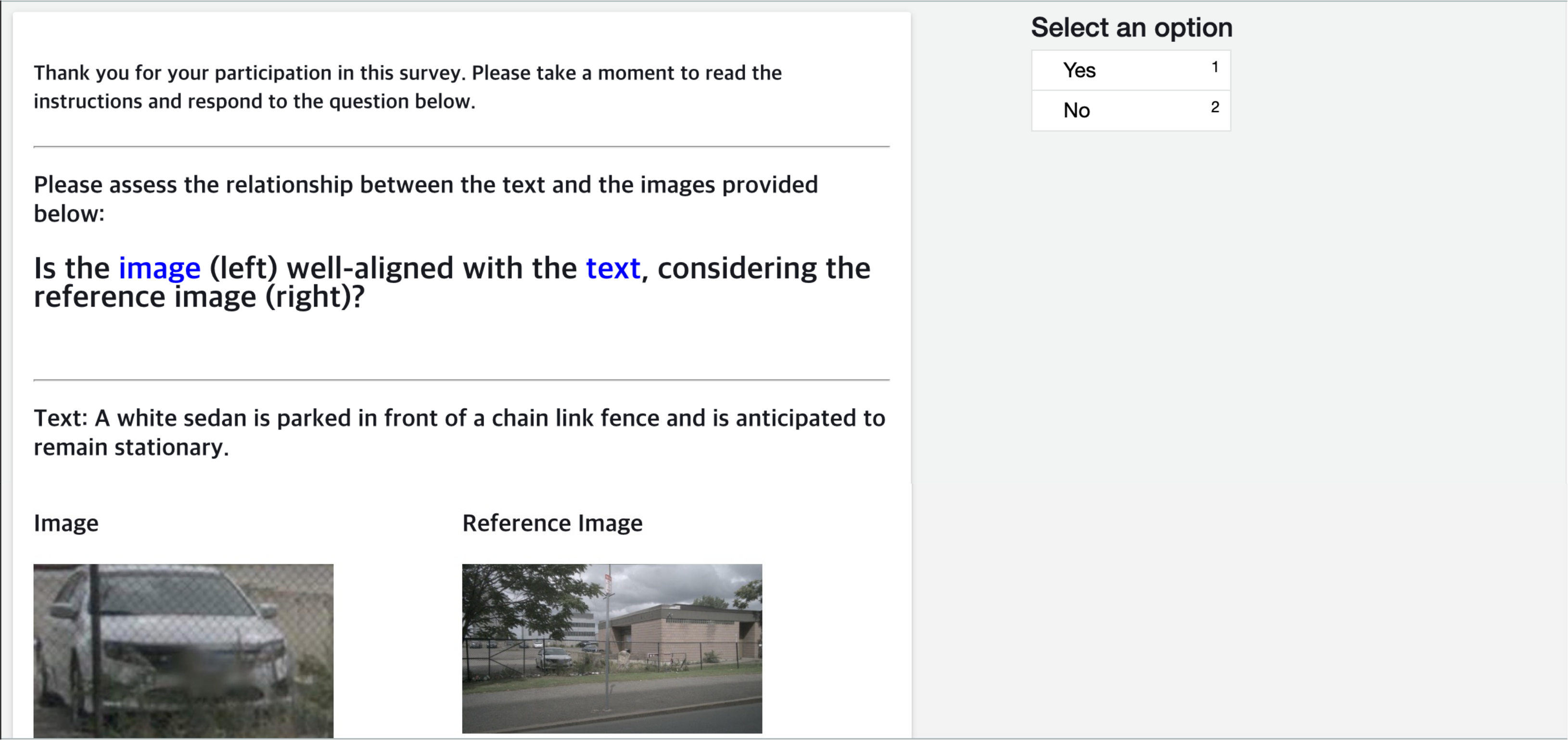}
    \vspace{-2em}
    \caption{Example of the Mechanical Turk evaluation interface used for assessing the alignment between generated text descriptions and corresponding images in the nuscenes-text dataset.}
    \label{fig:supple_mturk}
    \vspace{-2.em}
\end{figure*}
\myparagraph{More Details about Dataset Statistics.}
We further explore the details of the dataset we have created. The dataset contains 15,369,058 words, leading to a total of 17,134,981 tokens. This significant amount of text reflects the dataset's comprehensive scope, encompassing a variety of subjects and scenarios relevant to autonomous vehicles. With an average of 13.08 words and 14.58 tokens per text, the dataset showcases a wide-ranging vocabulary and linguistic diversity. Additionally, An example of the Mturk evaluation interface we used can be seen in Fig.~\ref{fig:supple_mturk}. The results from the human evaluation conducted via Mechanical Turk further demonstrate how well the captions included in our dataset describe the corresponding objects, indicating their substantial validity.

\myparagraph{More Details about nuScenes-Text Dataset.}
In this section, we provide additional examples of our created nuScenes-Text dataset. Fig.~\ref{fig:supple_dataset} represents textual descriptions obtained from surround-view images. Each agent has three distinct versions of textual descriptions and shows this descriptions of each agent in the bounding box.
The description generated through VLM in the top center image (CAM\_FRONT) includes location information based on the perspective of the ego vehicle (highlighted in red). However, this may differ from the perspective of other vehicles and pedestrians. Additionally, the location data highlighted in red in the top right image (CAM\_FRONT\_RIGHT) indicates the position of a person located on the left side of the image, but from the perspective of the autonomous vehicle, it may inaccurately depict the location (from the perspective of the autonomous vehicle, the person is positioned to the right). Such inaccuracies in image-based location data have the potential to compromise the trajectory prediction functionality of the model. This issue is addressed by removing incorrect information through LLM, and the improvements are clearly evident in the refined captions. Through this, we demonstrate the capability to generate accurate textual descriptions for all objects visible in surround-view images.

\begin{figure*}[t]
    \centering
    \includegraphics[width=.95\linewidth]{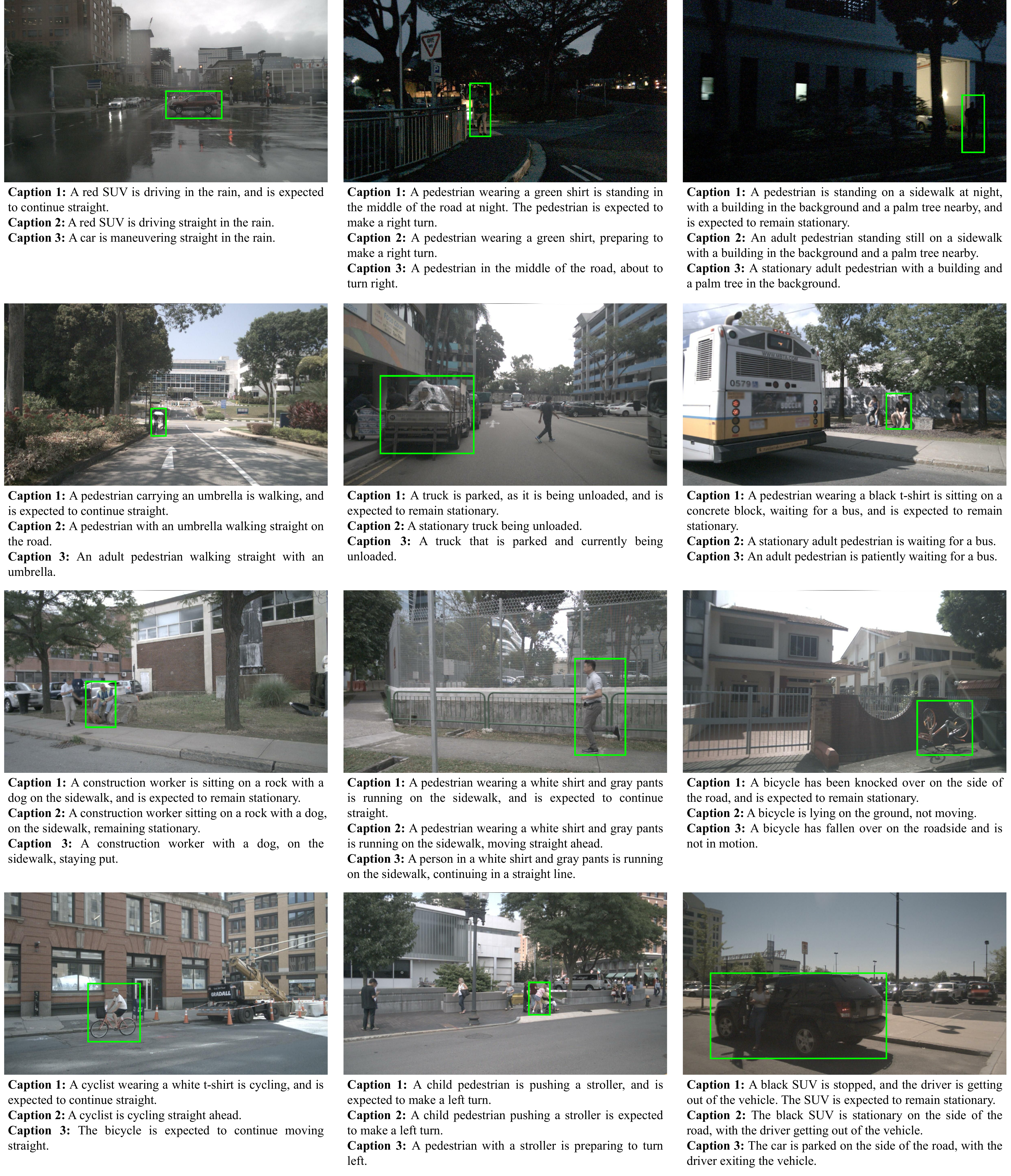}
    \vspace{-1em}
    \caption{Textual descriptions of unique scenarios in out dataset.}
    \label{fig:supple_dataset_unique}
    \vspace{-2em}
\end{figure*}

~\cref{fig:supple_dataset_unique} provides additional examples of unique situations that can be captured by camera images. Surprisingly, the textual description describes scenarios of rainy conditions and can also describe situations where camera data is compromised, such as low-light conditions. In addition, the text description shows that it can also capture situation information and details well, such as pedestrians holding umbrellas, unloading from trucks, people riding cycle, a driver getting out of a vehicle and pedestrians sitting on concrete blocks. Please refer to the images and captions together.

%% file: src_supple/sec/result.tex
\myparagraph{Additional Quantitative Results.}
In Table~\ref{tab:sup_table}, results for various agent types in the nuScenes whole dataset are presented. VisionTrap conducts predictions for both vehicles and pedestrians, showcasing information for both types. Model A employs only observed trajectories, Model B incorporates map data in addition to trajectory information, and Model C represents the results of VisionTrap. The outcomes demonstrate that the Visual Semantic Encoder and Text-driven Guidance Module contribute to improved performance across all agents.
{
\setlength{\tabcolsep}{2pt}
\renewcommand{\arraystretch}{1.3} 
\begin{table*}[t]
    \renewcommand{\arraystretch}{1.2}
    \caption{Results for various types: \textbf{A} uses only observed trajectory data, \textbf{B} without and \textbf{C} with our Visual Semantic Encoder and Text-driven Guidance Module. The data is used from the nuScenes~\cite{nuscenes} whole set.}
    \vspace{-.7em}
    \resizebox{.95\linewidth}{!}{%
        \begin{tabular}{c|ccc|ccc|ccc|>{\centering\arraybackslash}p{1.5cm}>{\centering\arraybackslash}p{1.5cm}>{\centering\arraybackslash}p{1.5cm}}
        \toprule
        \multirow{2}{*}{Model}  & \multicolumn{3}{c}{Type: All} & \multicolumn{3}{c}{Type: Vehicles} & \multicolumn{3}{c}{Type: Pedestrians} \\\cmidrule{2-4} \cmidrule{5-7} \cmidrule{8-10} 
        & ADE$_{10}$ $\downarrow$ & FDE$_{10}$ $\downarrow$ & MR$_{10}$ $\downarrow$ & ADE$_{10}$ $\downarrow$ & FDE$_{10}$ $\downarrow$ & MR$_{10}$ $\downarrow$ & ADE$_{10}$ $\downarrow$ & FDE$_{10}$ $\downarrow$ & MR$_{10}$ $\downarrow$ \\ \midrule\midrule

        A   & 0.425 &  0.641  &   0.081  & 0.453  &   0.702   & 0.102 & 0.341 & 0.471 & 0.019   \\        
        B   & 0.407  &  0.601  & 0.075  &   0.431   &   0.649  &  0.095  & 0.339 & 0.463 & 0.017  \\\midrule 
        C   & \textbf{0.368}  & \textbf{0.535} &  \textbf{0.051}    & \textbf{0.386} & \textbf{0.573} & \textbf{0.064} & \textbf{0.319}  & \textbf{0.430}  & \textbf{0.016}  \\ 
        \bottomrule
    \end{tabular}}
    
    \label{tab:sup_table}
    \vspace{-1.5em}
\end{table*}
}

\begin{figure*}[h]
    \centering
    \includegraphics[width=\linewidth]{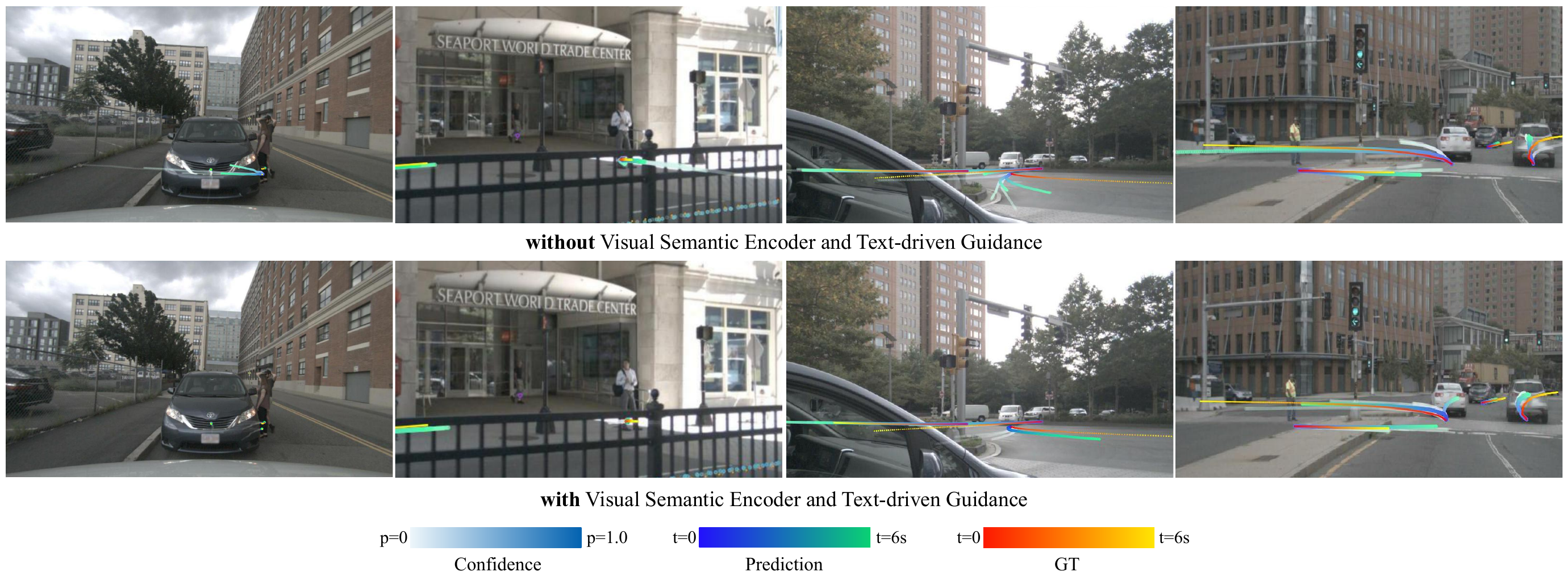}
    \caption{Examples where visual semantic information is used to improve the performance of trajectory prediction}
    \label{fig:supple_case_3-2}
    \vspace{-1.5em}
\end{figure*}

\begin{figure*}[t]
    \centering
    \includegraphics[width=\linewidth]{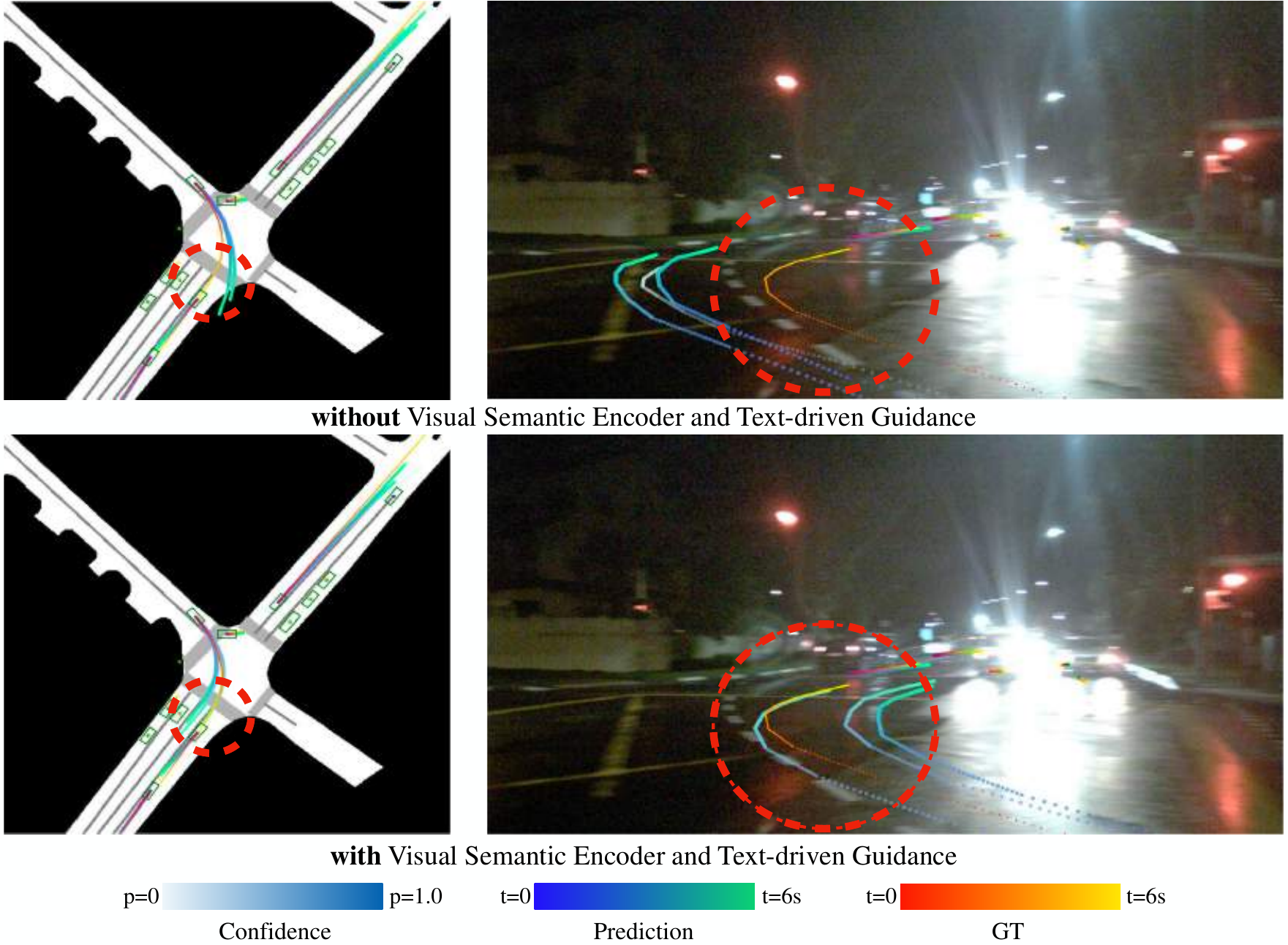}
    \vspace{-2em}
    \caption{Align trajectory to lane: Despite the use of nighttime images, it effectively aids in course adjustment when the vehicle makes a right turn.}
    \label{fig:supple_case_1-1}
    \vspace{-1.5em}
\end{figure*}

\begin{figure*}[t]
    \centering
    \includegraphics[width=\linewidth]{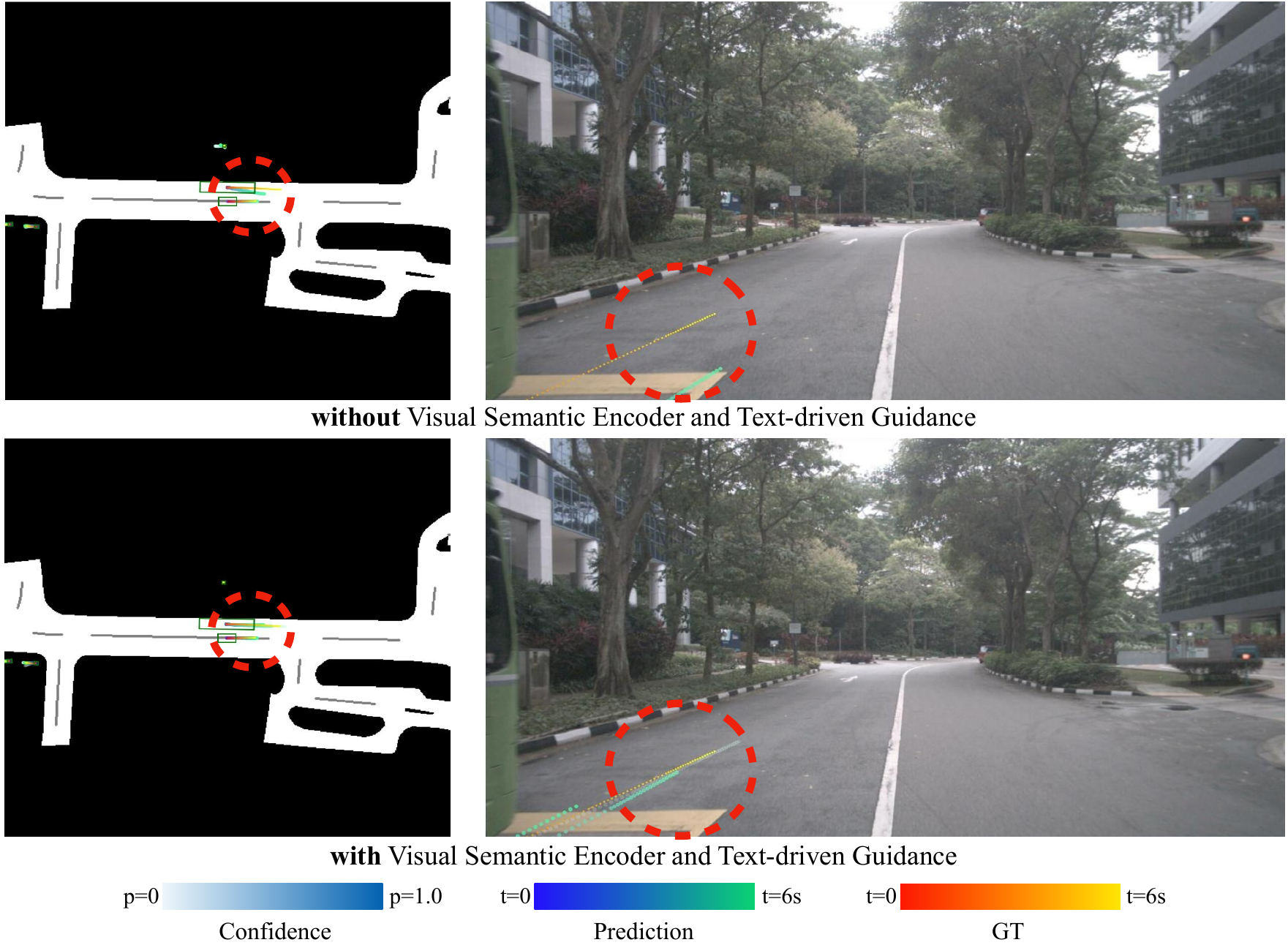}
    \vspace{-2em}
        \caption{Align trajectory to lane: The trajectory is adjusted to align with the lane when the parked bus starts moving.}
    \label{fig:supple_case_1-2}
    \vspace{-1.5em}
\end{figure*}

\begin{figure*}[t]
    \centering
    \includegraphics[width=\linewidth]{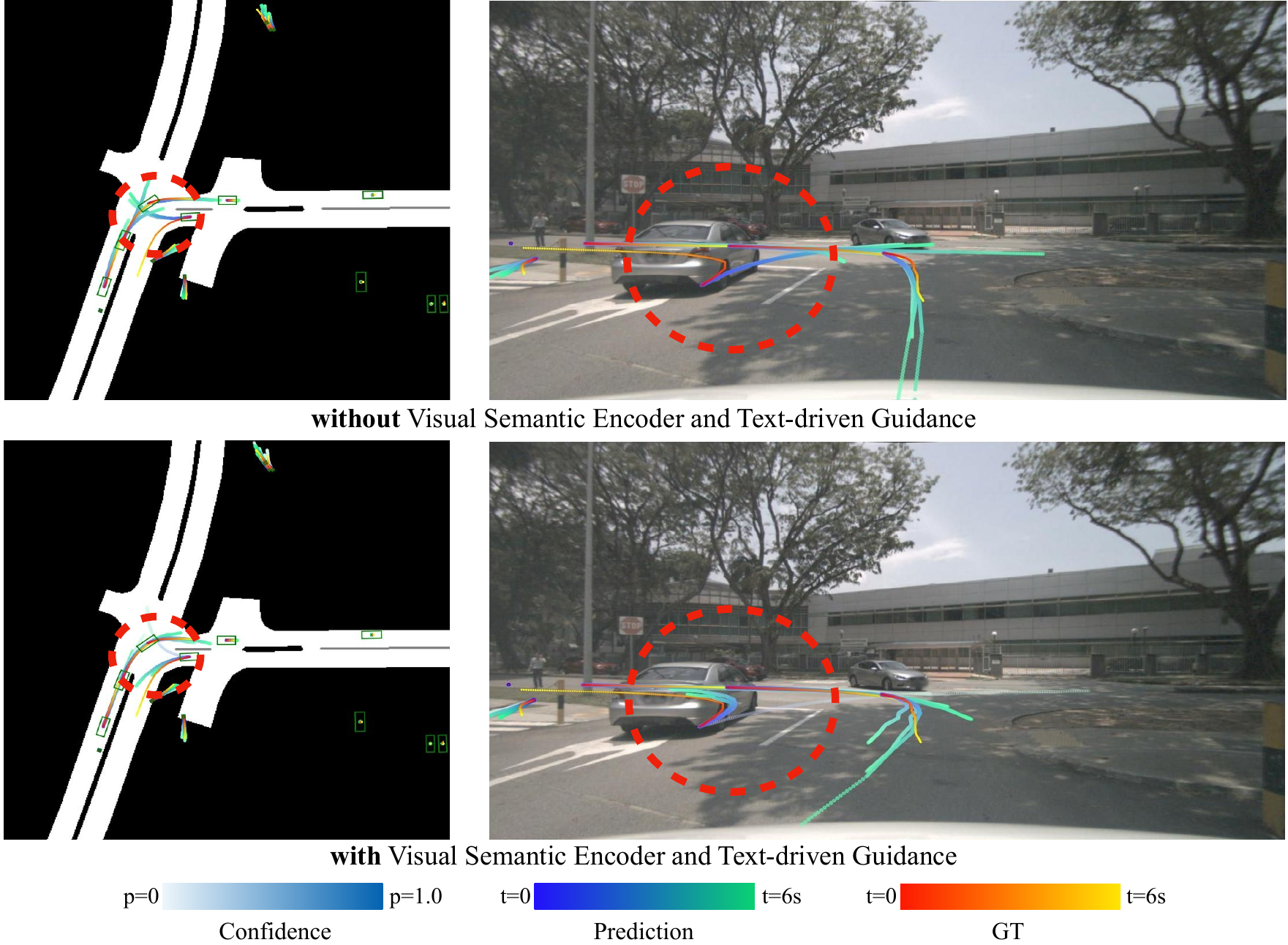}
    \vspace{-2em}
    \caption{Prevent collision: Vision data enables an understanding of the detailed situations of agents, enhancing interactions among them based on this understanding.}
    \label{fig:supple_case_2-1}
    \vspace{-1.5em}
\end{figure*}

\begin{figure*}[t]
    \centering
    \includegraphics[width=\linewidth]{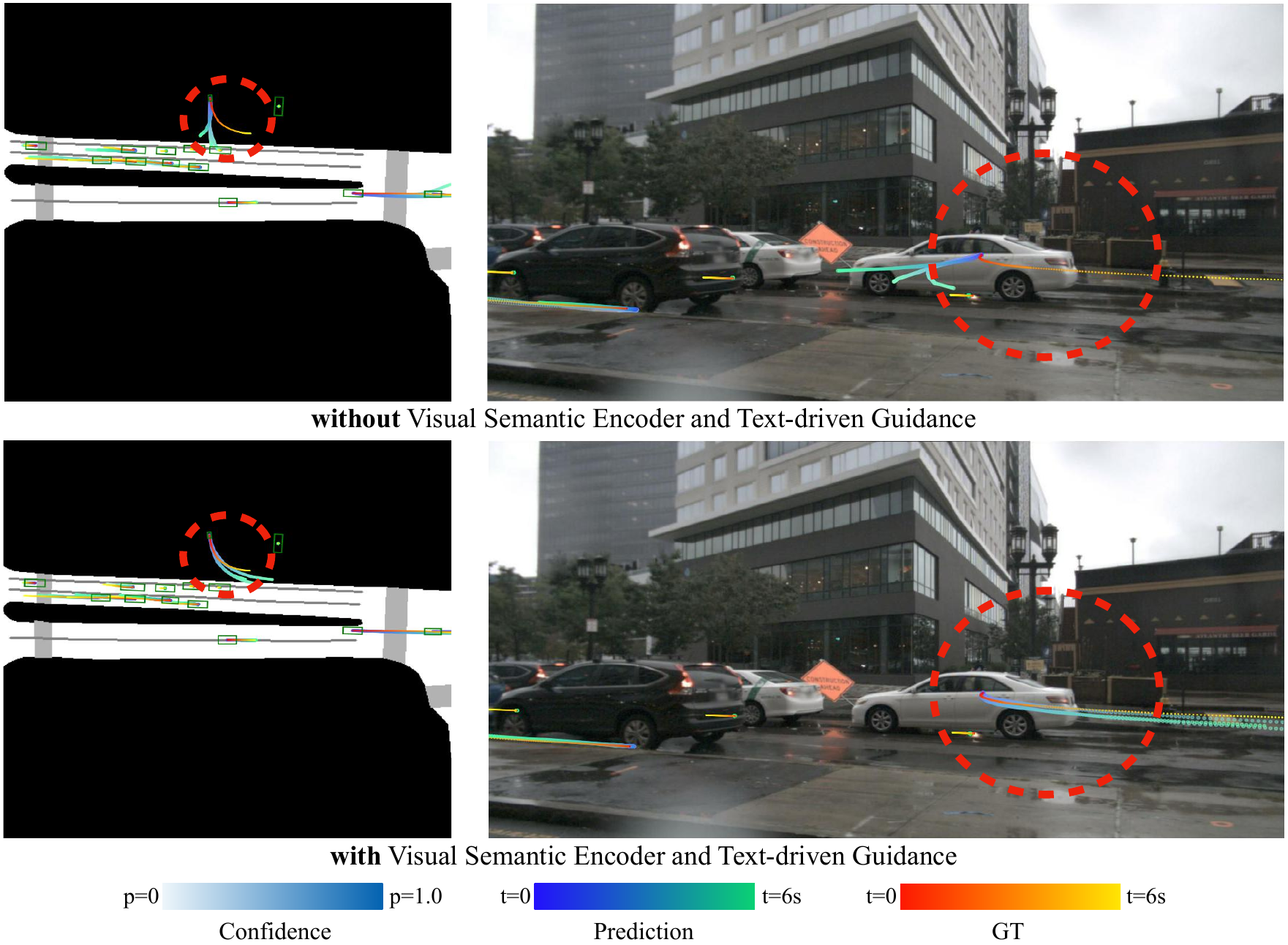}
    \vspace{-2.em}
    \caption{Prevent collision: The pedestrian's trajectory is adjusted to ensure there is no collision with the car and align with walking on the sidewalk.}
    \label{fig:supple_case_2-2}
    \vspace{-1.5em}
\end{figure*}

\begin{figure*}[t]
    \centering
    \includegraphics[width=\linewidth]{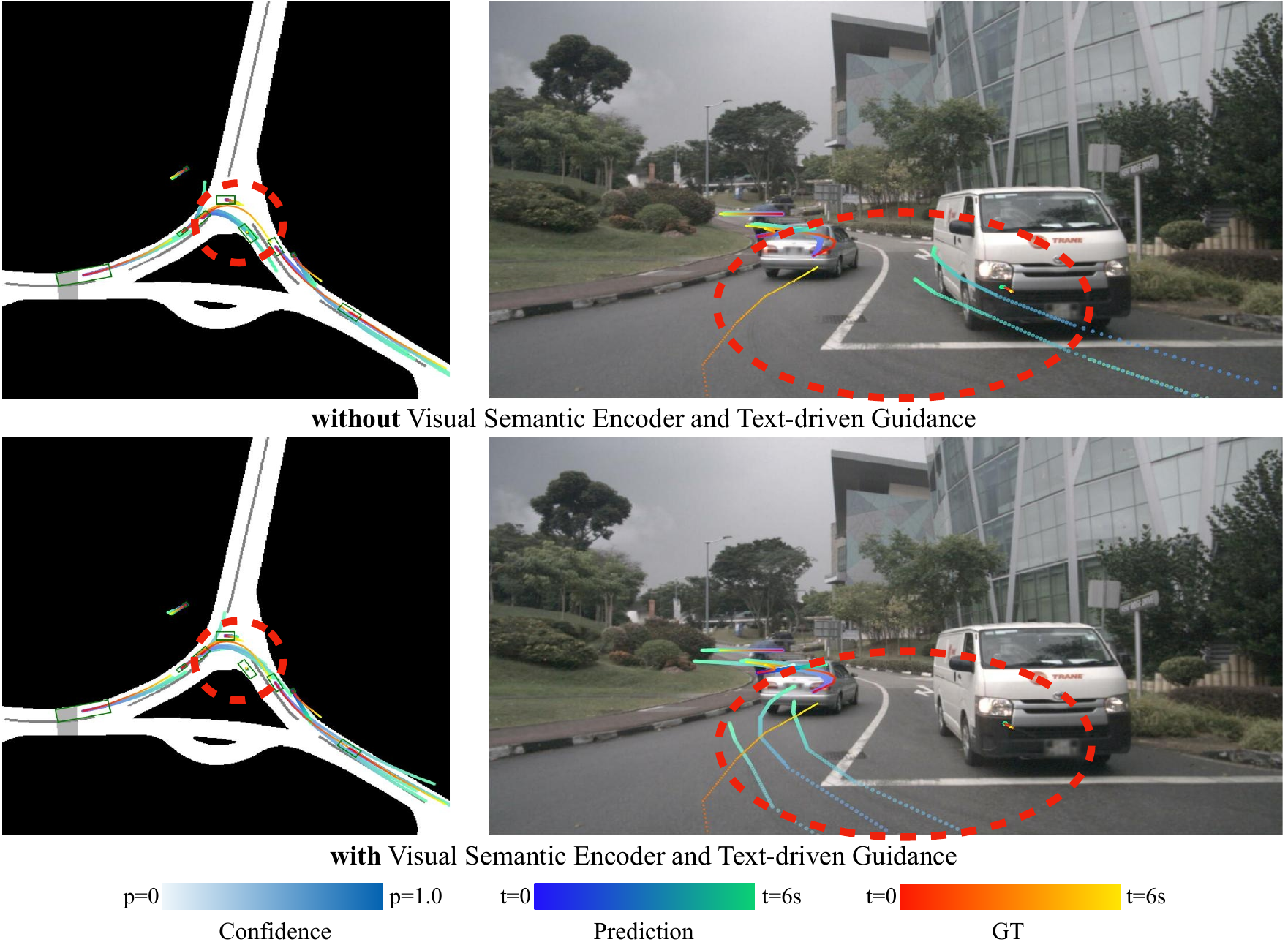}
    \vspace{-2em}
    \caption{The direct utilization of vision information: Vision information can determine the direction of the lane and the heading of the agents.}
    \label{fig:supple_case_3-1}
    \vspace{-1.5em}
\end{figure*}

\begin{figure*}[t]
    \centering
    \includegraphics[width=\linewidth]{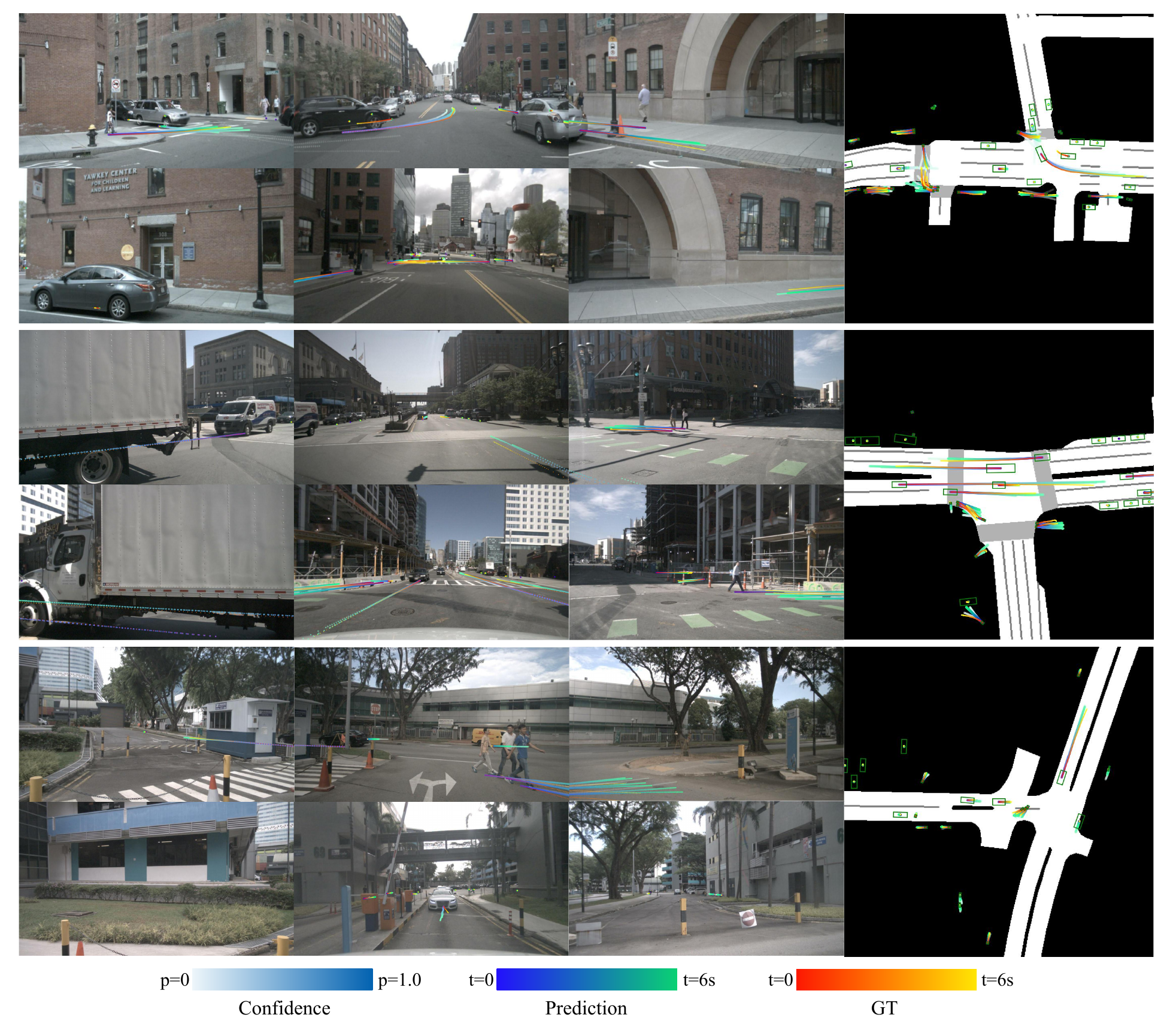}
    \vspace{-2em}
    \caption{Visualization results for trajectory prediction by our model for all objects (vehicles, pedestrians) in ego-centric surround view images.}
    \label{fig:supple_case_3-3}
    \vspace{-1.5em}
\end{figure*}

\myparagraph{Additional Qualitative Examples.}
We present additional qualitative examples obtained from various scenes. The examples are selected from the nuScenes dataset. Results from Fig.\ref{fig:supple_case_3-2} to Fig.\ref{fig:supple_case_3-3} illustrate without and with our Visual Semantic Encoder and Text-driven Guidance Module. Refer to the respective captions for explanations about the figures.




%% file: main.bbl
\begin{thebibliography}{10}
\providecommand{\url}[1]{\texttt{#1}}
\providecommand{\urlprefix}{URL }
\providecommand{\doi}[1]{https://doi.org/#1}

\bibitem{gpt3}
Brown, T., Mann, B., Ryder, N., Subbiah, M., Kaplan, J.D., Dhariwal, P., Neelakantan, A., Shyam, P., Sastry, G., Askell, A., et~al.: Language models are few-shot learners. Advances in neural information processing systems  \textbf{33},  1877--1901 (2020)

\bibitem{poly}
Buhet, T., Wirbel, E., Bursuc, A., Perrotton, X.: Plop: Probabilistic polynomial objects trajectory planning for autonomous driving. arXiv preprint arXiv:2003.08744  (2020)

\bibitem{nuscenes}
Caesar, H., Bankiti, V., Lang, A.H., Vora, S., Liong, V.E., Xu, Q., Krishnan, A., Pan, Y., Baldan, G., Beijbom, O.: nuscenes: A multimodal dataset for autonomous driving. In: Proceedings of the IEEE/CVF conference on computer vision and pattern recognition. pp. 11621--11631 (2020)

\bibitem{intentnet}
Casas, S., Luo, W., Urtasun, R.: Intentnet: Learning to predict intention from raw sensor data. In: Conference on Robot Learning. pp. 947--956. PMLR (2018)

\bibitem{multipath}
Chai, Y., Sapp, B., Bansal, M., Anguelov, D.: Multipath: Multiple probabilistic anchor trajectory hypotheses for behavior prediction. arXiv preprint arXiv:1910.05449  (2019)

\bibitem{debiasedcl}
Chuang, C.Y., Robinson, J., Lin, Y.C., Torralba, A., Jegelka, S.: Debiased contrastive learning. Advances in neural information processing systems  \textbf{33},  8765--8775 (2020)

\bibitem{p2t}
Deo, N., Trivedi, M.M.: Trajectory forecasts in unknown environments conditioned on grid-based plans. arXiv preprint arXiv:2001.00735  (2020)

\bibitem{pgp}
Deo, N., Wolff, E., Beijbom, O.: Multimodal trajectory prediction conditioned on lane-graph traversals. In: Conference on Robot Learning. pp. 203--212. PMLR (2022)

\bibitem{bert}
Devlin, J., Chang, M.W., Lee, K., Toutanova, K.: Bert: Pre-training of deep bidirectional transformers for language understanding. arXiv preprint arXiv:1810.04805  (2018)

\bibitem{tpnet}
Fang, L., Jiang, Q., Shi, J., Zhou, B.: Tpnet: Trajectory proposal network for motion prediction. In: Proceedings of the IEEE/CVF Conference on Computer Vision and Pattern Recognition. pp. 6797--6806 (2020)

\bibitem{vectornet}
Gao, J., Sun, C., Zhao, H., Shen, Y., Anguelov, D., Li, C., Schmid, C.: Vectornet: Encoding hd maps and agent dynamics from vectorized representation. In: Proceedings of the IEEE/CVF Conference on Computer Vision and Pattern Recognition. pp. 11525--11533 (2020)

\bibitem{HOME}
Gilles, T., Sabatini, S., Tsishkou, D., Stanciulescu, B., Moutarde, F.: Home: Heatmap output for future motion estimation. In: 2021 IEEE International Intelligent Transportation Systems Conference (ITSC). pp. 500--507. IEEE (2021)

\bibitem{thomas}
Gilles, T., Sabatini, S., Tsishkou, D., Stanciulescu, B., Moutarde, F.: Thomas: Trajectory heatmap output with learned multi-agent sampling. arXiv preprint arXiv:2110.06607  (2021)

\bibitem{gohome}
Gilles, T., Sabatini, S., Tsishkou, D., Stanciulescu, B., Moutarde, F.: Gohome: Graph-oriented heatmap output for future motion estimation. In: 2022 international conference on robotics and automation (ICRA). pp. 9107--9114. IEEE (2022)

\bibitem{loki}
Girase, H., Gang, H., Malla, S., Li, J., Kanehara, A., Mangalam, K., Choi, C.: Loki: Long term and key intentions for trajectory prediction. In: Proceedings of the IEEE/CVF International Conference on Computer Vision. pp. 9803--9812 (2021)

\bibitem{selecmix}
Hwang, I., Lee, S., Kwak, Y., Oh, S.J., Teney, D., Kim, J.H., Zhang, B.T.: Selecmix: Debiased learning by contradicting-pair sampling. Advances in Neural Information Processing Systems  \textbf{35},  14345--14357 (2022)

\bibitem{wcl}
Jang, T., Wang, X.: Difficulty-based sampling for debiased contrastive representation learning. In: Proceedings of the IEEE/CVF Conference on Computer Vision and Pattern Recognition (CVPR). pp. 24039--24048 (June 2023)

\bibitem{align}
Jia, C., Yang, Y., Xia, Y., Chen, Y.T., Parekh, Z., Pham, H., Le, Q., Sung, Y.H., Li, Z., Duerig, T.: Scaling up visual and vision-language representation learning with noisy text supervision. In: International conference on machine learning. pp. 4904--4916. PMLR (2021)

\bibitem{blip2}
Li, J., Li, D., Savarese, S., Hoi, S.: Blip-2: Bootstrapping language-image pre-training with frozen image encoders and large language models. arXiv preprint arXiv:2301.12597  (2023)

\bibitem{bevdepth}
Li, Y., Ge, Z., Yu, G., Yang, J., Wang, Z., Shi, Y., Sun, J., Li, Z.: Bevdepth: Acquisition of reliable depth for multi-view 3d object detection. In: Proceedings of the AAAI Conference on Artificial Intelligence. vol.~37, pp. 1477--1485 (2023)

\bibitem{lanegcn}
Liang, M., Yang, B., Hu, R., Chen, Y., Liao, R., Feng, S., Urtasun, R.: Learning lane graph representations for motion forecasting. In: Computer Vision--ECCV 2020: 16th European Conference, Glasgow, UK, August 23--28, 2020, Proceedings, Part II 16. pp. 541--556. Springer (2020)

\bibitem{fpn}
Lin, T.Y., Doll{\'a}r, P., Girshick, R., He, K., Hariharan, B., Belongie, S.: Feature pyramid networks for object detection. In: Proceedings of the IEEE conference on computer vision and pattern recognition. pp. 2117--2125 (2017)

\bibitem{spatiotemporal}
Liu, B., Adeli, E., Cao, Z., Lee, K.H., Shenoi, A., Gaidon, A., Niebles, J.C.: Spatiotemporal relationship reasoning for pedestrian intent prediction. IEEE Robotics and Automation Letters  \textbf{5}(2),  3485--3492 (2020)

\bibitem{laformer}
Liu, M., Cheng, H., Chen, L., Broszio, H., Li, J., Zhao, R., Sester, M., Yang, M.Y.: Laformer: Trajectory prediction for autonomous driving with lane-aware scene constraints. arXiv preprint arXiv:2302.13933  (2023)

\bibitem{mmtransformer}
Liu, Y., Zhang, J., Fang, L., Jiang, Q., Zhou, B.: Multimodal motion prediction with stacked transformers. In: Proceedings of the IEEE/CVF Conference on Computer Vision and Pattern Recognition. pp. 7577--7586 (2021)

\bibitem{cosschedule}
Loshchilov, I., Hutter, F.: Sgdr: Stochastic gradient descent with warm restarts. arXiv preprint arXiv:1608.03983  (2016)

\bibitem{adamw}
Loshchilov, I., Hutter, F.: Decoupled weight decay regularization. arXiv preprint arXiv:1711.05101  (2017)

\bibitem{drama}
Malla, S., Choi, C., Dwivedi, I., Choi, J.H., Li, J.: Drama: Joint risk localization and captioning in driving. In: Proceedings of the IEEE/CVF Winter Conference on Applications of Computer Vision. pp. 1043--1052 (2023)

\bibitem{titan}
Malla, S., Dariush, B., Choi, C.: Titan: Future forecast using action priors. In: Proceedings of the IEEE/CVF Conference on Computer Vision and Pattern Recognition. pp. 11186--11196 (2020)

\bibitem{umap}
McInnes, L., Healy, J., Melville, J.: Umap: Uniform manifold approximation and projection for dimension reduction. arXiv preprint arXiv:1802.03426  (2018)

\bibitem{mhajam}
Messaoud, K., Deo, N., Trivedi, M.M., Nashashibi, F.: Trajectory prediction for autonomous driving based on multi-head attention with joint agent-map representation (2020)

\bibitem{debcse}
Miao, P., Du, Z., Zhang, J.: Debcse: Rethinking unsupervised contrastive sentence embedding learning in the debiasing perspective. In: Proceedings of the 32nd ACM International Conference on Information and Knowledge Management. pp. 1847--1856 (2023)

\bibitem{wayformer}
Nayakanti, N., Al-Rfou, R., Zhou, A., Goel, K., Refaat, K.S., Sapp, B.: Wayformer: Motion forecasting via simple \& efficient attention networks. In: 2023 IEEE International Conference on Robotics and Automation (ICRA). pp. 2980--2987. IEEE (2023)

\bibitem{scenetransformer}
Ngiam, J., Caine, B., Vasudevan, V., Zhang, Z., Chiang, H.T.L., Ling, J., Roelofs, R., Bewley, A., Liu, C., Venugopal, A., et~al.: Scene transformer: A unified architecture for predicting multiple agent trajectories. arXiv preprint arXiv:2106.08417  (2021)

\bibitem{cpc}
Oord, A.v.d., Li, Y., Vinyals, O.: Representation learning with contrastive predictive coding. arXiv preprint arXiv:1807.03748  (2018)

\bibitem{frm}
Park, D., Ryu, H., Yang, Y., Cho, J., Kim, J., Yoon, K.J.: Leveraging future relationship reasoning for vehicle trajectory prediction. arXiv preprint arXiv:2305.14715  (2023)

\bibitem{covernet}
Phan-Minh, T., Grigore, E.C., Boulton, F.A., Beijbom, O., Wolff, E.M.: Covernet: Multimodal behavior prediction using trajectory sets. In: Proceedings of the IEEE/CVF conference on computer vision and pattern recognition. pp. 14074--14083 (2020)

\bibitem{clip}
Radford, A., Kim, J.W., Hallacy, C., Ramesh, A., Goh, G., Agarwal, S., Sastry, G., Askell, A., Mishkin, P., Clark, J., et~al.: Learning transferable visual models from natural language supervision. In: International conference on machine learning. pp. 8748--8763. PMLR (2021)

\bibitem{pie}
Rasouli, A., Kotseruba, I., Kunic, T., Tsotsos, J.K.: Pie: A large-scale dataset and models for pedestrian intention estimation and trajectory prediction. In: ICCV (2019)

\bibitem{bifold}
Rasouli, A., Rohani, M., Luo, J.: Bifold and semantic reasoning for pedestrian behavior prediction. In: Proceedings of the IEEE/CVF International Conference on Computer Vision (ICCV). pp. 15600--15610 (October 2021)

\bibitem{pepscenes}
Rasouli, A., Yau, T., Lakner, P., Malekmohammadi, S., Rohani, M., Luo, J.: Pepscenes: A novel dataset and baseline for pedestrian action prediction in 3d. arXiv preprint arXiv:2012.07773  (2020)

\bibitem{multimodal}
Rasouli, A., Yau, T., Rohani, M., Luo, J.: Multi-modal hybrid architecture for pedestrian action prediction. In: 2022 IEEE Intelligent Vehicles Symposium (IV). pp. 91--97. IEEE (2022)

\bibitem{fjmp}
Rowe, L., Ethier, M., Dykhne, E.H., Czarnecki, K.: Fjmp: Factorized joint multi-agent motion prediction over learned directed acyclic interaction graphs. In: Proceedings of the IEEE/CVF Conference on Computer Vision and Pattern Recognition. pp. 13745--13755 (2023)

\bibitem{trajectron++}
Salzmann, T., Ivanovic, B., Chakravarty, P., Pavone, M.: Trajectron++: Dynamically-feasible trajectory forecasting with heterogeneous data. In: Computer Vision--ECCV 2020: 16th European Conference, Glasgow, UK, August 23--28, 2020, Proceedings, Part XVIII 16. pp. 683--700. Springer (2020)

\bibitem{narrowing}
Su, D.A., Douillard, B., Al-Rfou, R., Park, C., Sapp, B.: Narrowing the coordinate-frame gap in behavior prediction models: Distillation for efficient and accurate scene-centric motion forecasting. In: 2022 International Conference on Robotics and Automation (ICRA). pp. 653--659. IEEE (2022)

\bibitem{multipath++}
Varadarajan, B., Hefny, A., Srivastava, A., Refaat, K.S., Nayakanti, N., Cornman, A., Chen, K., Douillard, B., Lam, C.P., Anguelov, D., et~al.: Multipath++: Efficient information fusion and trajectory aggregation for behavior prediction. In: 2022 International Conference on Robotics and Automation (ICRA). pp. 7814--7821. IEEE (2022)

\bibitem{air2}
Wu, D., Wu, Y.: Air$^2$ for interaction prediction. arXiv preprint arXiv:2111.08184  (2021)

\bibitem{mmclvis}
Yuan, X., Lin, Z., Kuen, J., Zhang, J., Wang, Y., Maire, M., Kale, A., Faieta, B.: Multimodal contrastive training for visual representation learning. In: Proceedings of the IEEE/CVF Conference on Computer Vision and Pattern Recognition. pp. 6995--7004 (2021)

\bibitem{agentformer}
Yuan, Y., Weng, X., Ou, Y., Kitani, K.: Agentformer: Agent-aware transformers for socio-temporal multi-agent forecasting. In: Proceedings of the IEEE/CVF International Conference on Computer Vision (ICCV) (2021)

\bibitem{lanercnn}
Zeng, W., Liang, M., Liao, R., Urtasun, R.: Lanercnn: Distributed representations for graph-centric motion forecasting. In: 2021 IEEE/RSJ International Conference on Intelligent Robots and Systems (IROS). pp. 532--539. IEEE (2021)

\bibitem{dclr}
Zhou, K., Zhang, B., Zhao, X., Wen, J.R.: Debiased contrastive learning of unsupervised sentence representations  (2022)

\bibitem{qcnet}
Zhou, Z., Wang, J., Li, Y.H., Huang, Y.K.: Query-centric trajectory prediction. In: Proceedings of the IEEE/CVF Conference on Computer Vision and Pattern Recognition. pp. 17863--17873 (2023)

\bibitem{hivt}
Zhou, Z., Ye, L., Wang, J., Wu, K., Lu, K.: Hivt: Hierarchical vector transformer for multi-agent motion prediction. In: Proceedings of the IEEE/CVF Conference on Computer Vision and Pattern Recognition. pp. 8823--8833 (2022)

\bibitem{deformdetr}
Zhu, X., Su, W., Lu, L., Li, B., Wang, X., Dai, J.: Deformable detr: Deformable transformers for end-to-end object detection. arXiv preprint arXiv:2010.04159  (2020)

\end{thebibliography}
